\def\year{2020}\relax
\documentclass[letterpaper]{article}
\usepackage{aaai20}
\usepackage{times}
\usepackage{helvet}
\usepackage{courier}
\usepackage{url}
\usepackage{graphicx}
\usepackage{amsfonts}
\usepackage{amsmath}

\usepackage{multirow}
\usepackage{threeparttable}
\usepackage{tabularx}
\usepackage{paralist}

\usepackage{bm}
\usepackage{bbm}

\usepackage{amsthm}
\usepackage{booktabs}       
\usepackage{amsfonts}       
\usepackage{nicefrac}       
\usepackage{microtype}      

\usepackage{siunitx}

\usepackage{mleftright}

\usepackage{subfig}

\usepackage{xspace}
\newcommand{\model}{\textsc{GraphSim}\xspace}

\newcommand{\nop}[1]{}

\newcommand*\samethanks[1][\value{footnote}]{\footnotemark[#1]}

\title{Learning-based Efficient Graph Similarity Computation via Multi-Scale Convolutional Set Matching}

\author{
  Yunsheng Bai,\thanks{The two first authors made equal contributions.}\textsuperscript{\rm 1}
  Hao Ding,\samethanks\textsuperscript{\rm 2}\thanks{This work is done before Hao Ding joined AWS AI Labs.}
  Ken Gu,\textsuperscript{\rm 1}
  Yizhou Sun,\textsuperscript{\rm 1}
  Wei Wang\textsuperscript{\rm 1}\\
  \textsuperscript{\rm 1}University of California, Los Angeles, \textsuperscript{\rm 2}AWS AI Labs\\
  yba@ucla.edu, haodin@amazon.com, ken.qgu@gmail.com, \{yzsun,weiwang\}@cs.ucla.edu \\
}

\begin{document}
\maketitle
\begin{abstract}
Graph similarity computation is one of the core operations in many graph-based applications, such as graph similarity search, graph database analysis, graph clustering, etc. Since computing the exact distance/similarity between two graphs is typically NP-hard, a series of approximate methods have been proposed with a trade-off between accuracy and speed. 
Recently, several data-driven approaches based on neural networks have been proposed, most of which model the graph-graph similarity as the inner product of their graph-level representations, with different techniques proposed for generating one embedding per graph. However, using one fixed-dimensional embedding per graph may fail to fully capture graphs in varying sizes and link structures---a limitation that is especially problematic for the task of graph similarity computation, where the goal is to find the fine-grained difference between two graphs. In this paper, we address the problem of graph similarity computation from another perspective, by directly matching two sets of node embeddings without the need to use fixed-dimensional vectors to represent whole graphs for their similarity computation. The model, \model, achieves the state-of-the-art performance on four real-world graph datasets under six out of eight settings (here we count a specific dataset and metric combination as one setting), compared to existing popular methods for approximate Graph Edit Distance (GED) and Maximum Common Subgraph (MCS) computation.
\end{abstract}

\section{Introduction}
\label{sec-intro}

Recent years we have witnessed the growing importance of graph-based applications in the domains of chemistry, bioinformatics, recommender systems, social network study, static program analysis, etc. One of the fundamental problems related to graphs is the computation of distance/similarity between two graphs. It not only is a core operation in graph similarity search and graph database analysis~\cite{zeng2009comparing,wang2012efficient}, but also plays a significant role in a wide range of applications. For example, in computer security, similarity between binary functions is useful for plagiarism and malware detection~\cite{xu2017neural}; in anomaly detection, similarity between communication graphs could help identify network intrusions from the graph-based connection records~\cite{noble2003graph}; in social network analysis, similarity between different user message graphs may reveal interesting behavioral patterns ~\cite{koutra2013deltacon}.

Among various definitions of graph similarity/distance, Graph Edit Distance (GED)~\cite{bunke1983distance} and Maximum Common Subgraph (MCS)~\cite{bunke1998graph} are two popular and domain-agnostic metrics. 
However, the computation of exact GED and MCS is known to be NP-hard~\cite{zeng2009comparing,bunke1998graph}, incurring significant computational burden in practice~\cite{blumenthal2018exact}. For example, a recent study shows that even the state-of-the-art algorithms cannot reliably compute the exact GED  between graphs of more than 16 nodes within a reasonable time~\cite{blumenthal2018exact}.

Given the great significance yet huge challenge of computing the exact graph distance/similarity, various approximate algorithms have been proposed to compute the graph distance/similarity in a fast but heuristic way, including traditional algorithmic approaches~\cite{riesen2009approximate,fankhauser2011speeding,daller2018approximate} as well as more recent data-driven neural network approaches~\cite{ribalearning,ktena2017distance,bai2018graph,li2019graph}.

Compared with traditional algorithmic approaches which typically involve knowledge and heuristics specific to a metric, the neural network approaches \textit{learn} graph similarity from data: During training, the parameters are learned by minimizing the loss between the predicted similarity scores and the ground truth; during testing, unseen pairs of graphs can be fed into these models for fast approximation of their similarities.

However, a major limitation of most current neural network models is that they rely on graph-level embeddings to model the similarity of graphs: Each graph is first represented as a fixed-length vector, and then the similarity of two graphs can be modeled as a vector operation on the two embeddings, e.g. cosine similarity. 
However, real-world graphs typically come in very different sizes, which the fixed-length vector representation may fail to fully capture. Even when the two graphs of interest are similar in sizes, the actual difference between them can lie in very small local substructures, which is hard to be captured by the single vector. This is especially problematic for the task of graph similarity computation, where the goal is to compare the difference between all nodes and edges of the two graphs. 
For simple or regular graphs, this approach may work well, but for more complicated scenarios in which graphs are of very different structures and/or the task is to find the fine-grained node-node correspondence~\cite{zanfir2018deep}, this approach often produces less effective models.


In this paper, we propose to avoid the generation of graph-level embeddings, and instead directly perform neural operations on the two sets of node embeddings. Inspired by two classic families of algorithms for graph similarity/distance~\cite{nikolentzos2017matching,riesen2009approximate,fankhauser2011speeding}, our model \model turns the two sets of node embeddings into a similarity matrix consisting of the pairwise node-node similarity scores, and is trained in an end-to-end fashion (Fig.~\ref{fig:model}). By carefully ordering the nodes in each graph,
the similarity matrix encodes the similarity patterns specific to the graph pair, which allows the standard image processing techniques to be adapted to model the graph-graph similarity. The new challenges in the graph setting compared to the standard image processing using Convolutional Neural Networks (CNN) are that:

\begin{itemize}
\item\textit{\textbf{Permutation invariance}}. The same graph can be represented by different adjacency matrices by permuting the order of nodes, and the model should not be sensitive to such permutation. 
\item \textit{\textbf{Spatial locality preservation}}. CNN architectures assume the input data has spatial locality, i.e, close-by data points are more similar to each other. How to make our embedding-based similarity matrix preserve such spatial locality is important.
\item\textit{\textbf{Graph size invariance}}. The CNN architecture requires fixed-length input. How to handle graphs with different sizes is another question to address.
\item \textit{\textbf{Multi-scale comparison.}} Finally, graphs naturally contain patterns of different scales that may be unknown in advance. The algorithm should be able to capture and leverage structural information and features of multiple granularities.
\end{itemize}

To tackle these challenges, we propose \model, which addresses the graph similarity computation task in a novel way by direct usage of node-level embeddings via pairwise node-node similarity scores. We show that \model can be combined with various node embedding approaches, improving performance on four graph similarity computation datasets under six out of eight settings.
Finally, we show that \model can learn interpretable similarity patterns that exist in the input graph pairs.

\section{Related Work}
\label{sec-related}


\textbf{Graph Representation Learning} \enspace Over the years, there have been a great number of works dealing with the representation of nodes~\cite{hamilton2017inductive}, and graphs~\cite{ying2018hierarchical}. Among the node embedding methods, neighbor aggregation based methods, e.g. GCN~\cite{kipf2016semi}, GraphSAGE~\cite{hamilton2017inductive},
GIN~\cite{xu2018powerful}, etc., are permutation-invariant, and have gained a lot of attention. 

Neural network based methods have been used in a broad range of graph applications, most of which are framed as node-level prediction tasks~\cite{hamilton2018querying} or single graph classification~\cite{ying2018hierarchical}. In this work, we consider the task of graph similarity computation, which is under the general problem of graph matching~\cite{emmert2016fifty}.

\textbf{Text and Graph Matching with Neural Networks} \enspace Text matching has a long history with many successful applications~\cite{mitra2017learning}. Among various methods for text matching, promising results in matching sequences of word embeddings~\cite{he2016pairwise}
inspire us to explore the potential of using node embeddings for the task of graph matching directly without graph-level representations. In contrast, neural network based graph matching remains largely unexplored, and most existing works still rely on first generating one embedding per graph using graph neural networks, and then modeling the graph-graph similarity using the two graph-level representations. 

We examine several existing works on similarity computation for graphs: (1) \textsc{Siamese MPNN} (\textsc{SMPNN})~\cite{ribalearning} is an early work that models the similarity as a simple summation of certain node-node similarity scores. (2) \textsc{GCNMean} and \textsc{GCNMax}~\cite{ktena2017distance} apply the GCN architectures with graph coarsening~\cite{defferrard2016convolutional} to generate graph-level embeddings for the similarity.
(3) \textsc{SimGNN}~\cite{bai2018graph} attempts to use node-node similarity scores by taking their histogram features, but still largely relies on the graph-level embeddings due to the histogram function being non-differentiable. (4) \textsc{GMN}~\cite{li2019graph} is a recent work which manages to introduce node-node similarity information into graph-level embeddings via a cross-graph attention mechanism, but the cross-graph communication only updates the node embeddings, and still generates one embedding per graph from the updated node embeddings.

\section{Problem Definition}
\label{sec-prelim}

Graphs are data structures with a node set $\mathcal{V}$ and a edge set $\mathcal{E}$,  $\mathcal{G}=(\mathcal{V},\mathcal{E})$, where $\mathcal{E}\subseteq \mathcal{V} \times \mathcal{V}$. The number of nodes of $\mathcal{V}$ is denoted as $N=|\mathcal{V}|$. Each node and edge can be associated with labels, such as atom and chemical bond type in a molecular graph. In this study, we confine our graphs as undirected and unweighted graphs, but it is not hard to extend \model to other types of graphs, since \model is a general framework for graph similarity computation.





Given two graphs $\mathcal{G}_1$ and $\mathcal{G}_2$, different distance/similarity metrics can be defined. 

\textbf{Graph Edit Distance (GED)} \enspace  The edit distance between two graphs $\mathcal{G}_1$ and $\mathcal{G}_2$
is the number of edit operations in the optimal alignments that transform $\mathcal{G}_1$ into $\mathcal{G}_2$, where an edit operation on a graph $\mathcal{G}$ is an insertion or deletion of a node/edge or relabelling of a node~\footnote{Other variants of GED definitions exist~\cite{riesen2013novel}, and we adopt this basic version for each in this work.}.
We transform GED into a similarity metric ranging between 0 and 1 using a one-to-one mapping function.

\textbf{Maximum Common Subgraph (MCS)} \enspace A Maximum Common Subgraph of two graphs $\mathcal{G}_1$ and $\mathcal{G}_2$
is a subgraph common to both $\mathcal{G}_1$ and $\mathcal{G}_2$ such that there is no other subgraph of $\mathcal{G}_1$ and $\mathcal{G}_2$ with more nodes.
The MCS definitions have two variants~\cite{raymond2002maximum}: In one definition, the MCS must be a connected graph; in the other, the MCS can be disconnected. In this paper, we adopt the former definition. For $\mathcal{G}_1$ and $\mathcal{G}_2$, their similarity score is defined as the number of nodes in their MCS, i.e. $|\mathrm{MCS}(\mathcal{G}_1,\mathcal{G}_2)|$.


\textbf{Learning-based Graph Similarity Computation} \enspace 
Given a graph similarity definition, our goal is to learn a neural network based function that takes two graphs as input and outputs the desired similarity score through training, which can be applied to any unseen graphs for similarity computation at the test stage.  

In later sections, we will introduce how to design a neural network architecture to serve this purpose, and why such design is reasonable by providing  connections to set matching.

\nop{Our goal is to learn a neural network based function that takes two graphs as input and outputs the similarity score that can be transformed back to GED through a one-to-one mapping.
}

 \nop{We are given an undirected, unweighted graph $\mathcal{G}=(\mathcal{V},\mathcal{E})$ with $N=|\mathcal{V}|$ nodes. Node features are summarized in an $N\times D$ matrix $\bm{H}$. We transform GED into a similarity metric ranging between 0 and 1 via a one-to-one function (more details can be found in Section~\ref{subsec-data-preproc}). Our goal is to learn a neural network based function that takes two graphs as input and outputs the similarity score that can be transformed back to GED through the one-to-one mapping.}

\begin{figure}
\centering
\includegraphics[width=.95\columnwidth]{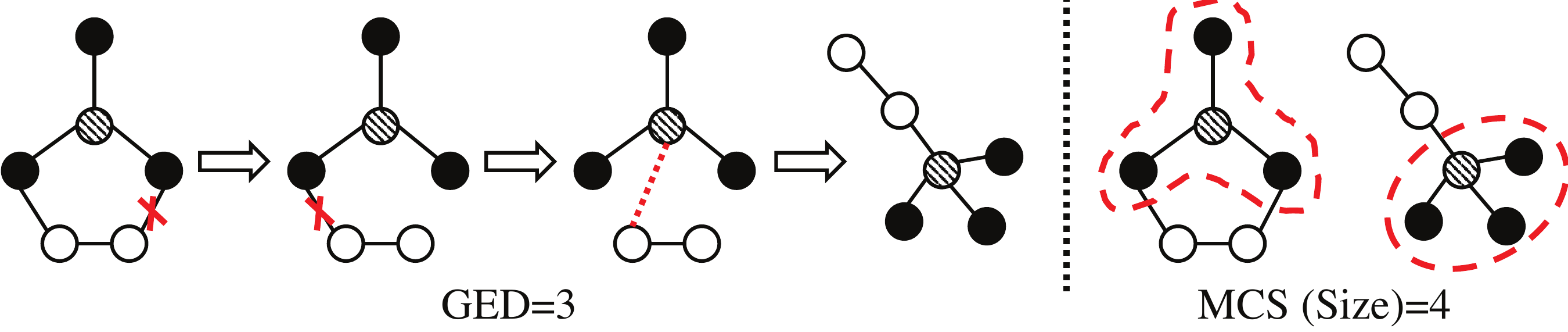}
\caption{The GED is 3, as the transformation needs 3 edit operations: Two edge deletions, and an edge insertion. The MCS size is 4. }
\label{fig:ged}
\end{figure}

\section{The Proposed Approach: {\model}} 
\label{sec-model}

\begin{figure*}[h]
\centering
\includegraphics[width=.95\textwidth]{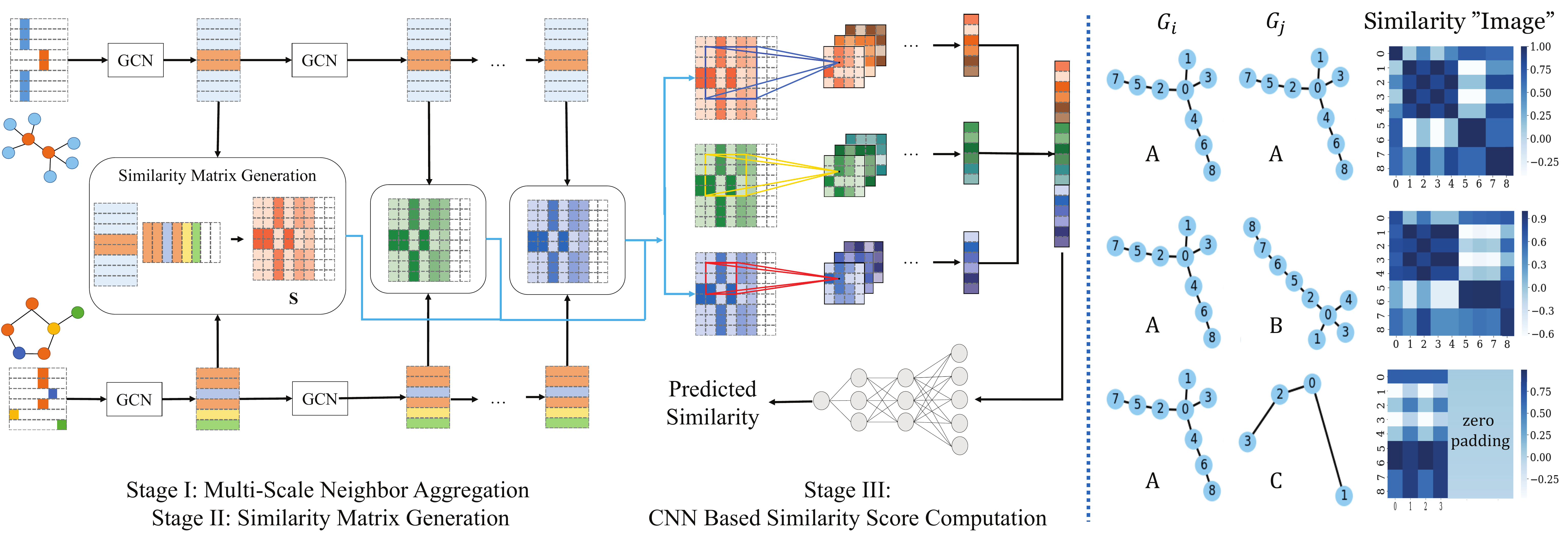}
\caption{Left: An overview illustration of our proposed method \model. No graph-level representation is generated, and it directly uses the node-node similarity scores in the three similarity matrices corresponding to node embeddings at different scales. Right: Illustration of three similarity matrices from the \textsc{Linux}~\cite{wang2012efficient} dataset. For two isomorphic graphs ($A$ and $A$), a strong symmetric diagonal block pattern is observed;
for two less similar graphs ($A$ and $B$), the dark diagonal pattern is less evident; for two graphs that are not similar at all ($A$ and $C$), the symmetric block pattern is almost gone. 
For graphs of different sizes, we devise a consistent max padding scheme. 
Intuitively, the block patterns can be thought of as graph-graph similarity patterns at different scales, which are suitable for CNNs to capture.}
\label{fig:model}
\end{figure*}

\model consists of the following sequential stages: 1) \textit{Multi-scale neighbor aggregation layers} generate vector representations for each node in the two graphs at different scales; 2) \textit{Similarity matrix generation layers} compute the inner products between the embeddings of every pair of nodes in the two graphs, resulting in multiple similarity matrices capturing the node-node interaction scores at different scales; 3) \textit{CNN layers} convert the similarity computation problem into a pattern recognition problem, which provides multi-scale features to a \textit{fully connected network} to obtain a final predicted graph-graph similarity score. An overview of our model is illustrated in Fig.~\ref{fig:model}.

\subsection{Multi-Scale Neighbor Aggregation} 
We build upon an active line of research on graph neural networks for generating node representations. Graph Convolutional Networks (GCN)~\cite{defferrard2016convolutional,kipf2016semi}, for example,
is a neighbor aggregation approach which generates node embeddings from the local substructure information of each node, 
which is an inductive method and can be applied to unseen nodes. In Fig.~\ref{fig:model}, different node types are represented by different colors and one-hot encoded as the initial node representation. For graphs with unlabeled nodes, we use the same constant vector as the initial representation. 

The core operation, graph convolution, operates on the representation of a node, which is denoted as $\bm{u_i} \in \mathbb{R}^{D}$, and is defined as follows:
\begin{equation} \label{eq:gcn} \mathrm{conv}(\bm{u_i}) = \mathrm{ReLU}( \sum_{j \in \mathcal{N}(i)} \frac{1}{\sqrt[]{d_i d_j}} \bm{u}_j \bm{W}^{(n)} + \bm{b}^{(n)})\end{equation} 
where $\mathcal{N}(i)$ is the set of the first-order neighbors of node $i$ plus $i$ itself, $d_i$ is the degree of node $i$ plus 1, $\bm{W}^{(n)} \in \mathbb{R}^{D^{(n)} \times D^{(n+1)}}$ is the weight matrix of the $n$-th GCN layer, $\bm{b}^{(n)} \in \mathbb{R}^{D^{(n+1)}}$ is the bias, $D^{(n)}$ denotes the dimensionality of embedding vector at layer $n$, and $\mathrm{ReLU}(x)=\mathrm{max}(0,x)$ is the activation function. 

Intuitively, the graph convolution operation aggregates the features from the first-order neighbors of the node. Since applying the GCN layer once on a node can be regarded as aggregating the representations of its first-order neighbors and itself, sequentially stacking $L$ layers would cause the final representation of a node to include its $L$-th order neighbors. In other words, the more GCN layers, the larger the scale of the learned embeddings. 

\textbf{Multi-Scale GCN} \enspace 
The potential issue of using a deep GCN architecture is that the embeddings may lose subtle patterns in local neighborhood after aggregating neighbors multiple times. The issue is especially severe when the two graphs are very similar, and the differences mainly lie in small local substructures.



One natural way for humans to compare the difference between two graphs is to recursively break down the whole graph into its compositional subgraphs via a top-down approach. Each subgraph is further decomposed into additional subgraph levels, until the entire specification is reduced to node level, producing a hierarchy of subgraph (de)composition. We therefore propose a multi-scale framework to extract the output of each of the many GCN layers for the construction of similarity matrices, which is shown next.


\model is a general framework for similarity computation, and can work with GCN and any of its successors such as  \textsc{GraphSAGE}~\cite{hamilton2017inductive}, 
\textsc{GIN}~\cite{xu2018powerful}, etc. 

\subsection{Similarity Matrix Generation}
\label{subsec-smg}


The use of node-node similarity scores roots in some classic approaches to modeling graph similarity, such as the Optimal Assignment Kernels for graph classification and the Bipartite Graph Matching for GED approximation. The detailed connection between the proposed model and these methods is in the supplementary material. Here we present some key properties of these methods that are needed to motivate the usage of node-node similarity scores.

\textbf{Optimal Assignment Kernels} \enspace Given two graphs with node embeddings $\bm{X} \in \mathbb{R}^{N_1 \times D}$ and $\bm{Y} \in \mathbb{R}^{N_2 \times D}$, the Earth Mover's Distance graph kernel~\cite{nikolentzos2017matching} solves the following transportation problem~\cite{rubner2000earth}:
\begin{equation}
\label{eq:mne}
\begin{aligned}
\min & \sum_{i=1}^{N_1} \sum_{j=1}^{N_2} \bm{T}_{ij} {||\bm{x}_i - \bm{y}_j||}_2 \\
\end{aligned}
\end{equation}
subject to several constraints. Intuitively, each of the two graphs is represented as a set of node embeddings, and the kernel finds the optimal way to transform one set of node embeddings to the other, with the cost to transform one pair of nodes being their Euclidean distance. 

\textbf{Bipartite Graph Matching} \enspace The \textsc{Hungarian}~\cite{riesen2009approximate} and \textsc{VJ}~\cite{fankhauser2011speeding} algorithms for approximate GED computation solve the following assignment problem form:
\begin{equation}
\label{eq:bgm}
\begin{aligned}
\min & \sum_{i=1}^{N^{'}} \sum_{j=1}^{N^{'}} \bm{T}_{ij} \bm{C}_{ij} \\
\end{aligned}
\end{equation}
subject to several constraints with cost matrix $\bm{C}$ denoting the insertion, deletion, and substitution costs associated with the GED metric for every node pair in the two graphs.

Despite different goals, both the kernel method and GED approximate algorithms work directly on node-level information without the whole-graph representations. Specifically, both need the pairwise node-node distance scores between the two graphs, with different definitions of node-node distances. 

Since \model is trained end-to-end for graph similarity computation, we can calculate the inner products between all pairs of node embeddings in the two graphs at multiple scales, resulting in multiple similarity matrices. Treat each matrix as an image, the task of graph similarity measurement is viewed as an image processing problem in which the goal is to discover the optimal node matching pattern encoded in the image by applying CNNs. Consider this process as a similarity operator that transforms two sets of node embeddings into a score. Then \model can be regarded as:
\begin{equation}
\label{eq:mne_equivalent}
\begin{aligned}
\min (h_{\bm{\Theta}} (\bm{X}, \bm{Y}) - s_{ij})^2
\end{aligned}
\end{equation}
where $h_{\Theta} (\bm{X}, \bm{Y})$ denotes the similarity matrix generation and its subsequent layers. The training is guided by the true similarity score $s_{ij}$ for the update of weights $\bm{\Theta}$ associated with the neighbor aggregation layers that generate node embeddings $\bm{X}$ and $\bm{Y}$ as well as the subsequent CNNs. In contrast, for the optimization problems~(\ref{eq:mne}) and (\ref{eq:bgm}), in order to find the optimal value as the computed graph distance, the problem must be solved explicitly~\cite{kuhn1955hungarian,jonker1987shortest}.



\textbf{BFS Ordering} \enspace 
Different from pixels of images or words of sentences, nodes of a graph typically lack ordering. A different node ordering would lead to a different similarity matrix. 
To completely solve the graph node permutation problem, we need to find one canonical ordering for each graph in a collection of graphs, which is NP-hard as shown in an early work on CNN for graphs~\cite{niepert2016learning}.
Moreover, the CNNs require spatial locality preservation. To alleviate these two issues, we utilize the breadth-first-search (BFS) node-ordering scheme proposed in GraphRNN~\cite{you2018graphrnn} to sort and reorder the node embeddings. Since BFS is performed on the graph, nearby nodes are ordered close to each other. 
It is worth noting that the BFS ordering scheme achieves a reasonable trade-off between efficiency and uniqueness of ordering, as it requires quadratic operations in the worst case (i.e. complete graphs)~\cite{you2018graphrnn}. 

Besides the issues related to node permutation and spatial locality, one must address the challenge posed by graph size variance and the fact that CNN architecture requires fix-length
input. To preserve the information of graph size variance while fix the size of similarity matrix, we propose the solutions as follows:

 


\textbf{Max Padding} \enspace 
One can fix the number of nodes in each graph by adding fake nodes to a pre-defined number, and therefore lead to a fix-size similarity matrix. However, such matrix will completely ignore the graph size information, which is pivotal as seen in the definitions of both GED and MCS. For example, the similarity matrix between two small but isomorphic graphs may be padded with a lot of zeros, potentially misleading the CNNs to predict they are dissimilar.

To reflect the difference in graph sizes in the similarity matrix, we devise max padding.
Suppose $\mathcal{G}_1$ and $\mathcal{G}_2$ contain $N_1$ and $N_2$ nodes respectively, we pad $|N_1 - N_2|$ rows of zeros to the node embedding matrix of the smaller of the two graphs, so that both graphs contain $\mathrm{max}(N_1,N_2)$ nodes. 


\textbf{Matrix Resizing} \enspace
To apply CNNs to the similarity matrices, 
We resize the simlarity matrices through image resampling~\cite{thevenaz2000image}. For implementation, we choose the bilinear interpolation, and leave the exploration of more advanced techniques for future work. The resulting similarity matrix $\bm{S}$ has fixed shape $M \times M$, where $M$ is a hyperparameter controlling the degree of information loss caused by resampling.


The following equation summarizes the similarity matrix generation process:
\begin{equation} \label{eq:gil} \bm{S} = \mathrm{RES}_{M}(\widetilde{\bm{H}}_1 \widetilde{\bm{H}}_2^{T})\end{equation}
where $\widetilde{\bm{H}_i} \in \mathbb{R}^{\mathrm{max}(N_1,N_2) \times D}, i \in \{1,2\}$ is the padded node embedding matrix $\bm{H}_i \in \mathbb{R}^{N_i \times D}, i \in \{1,2\}$ with zero or $|N_1 - N_2|$ nodes padded, and $\mathrm{RES}(\cdot): \mathbb{R}^{\mathrm{max}(N_1,N_2) \times \mathrm{max}(N_1,N_2)} \mapsto \mathbb{R}^{M \times M}$ is the resizing function, where $M$ is a hyperparameter controlling the degree of information loss caused by resampling.




\subsection{CNN Based Similarity Score Computation}
\label{subsec-cnn}




We feed these matrices through multiple CNNs in parallel. As shown in Fig.~\ref{fig:model}, three CNN ``channels'' are used, each with its own CNN filters. At the end, the results are concatenated and fed into multiple fully connected layers, so that a final similarity score $\hat{s}_{ij}$ is generated for the graph pair $\mathcal{G}_i$ and $\mathcal{G}_j$. The mean square error loss function is used to train our model: $ \mathcal{L} = \frac{1}{|\mathcal{D}|}\sum_{(i,j) \in \mathcal{D}} (\hat{s_{ij}} - s_{ij})^{2}$
where $\mathcal{D}$ is the set of training graph pairs, and $s_{ij}$ is the true similarity score coming from any graph similarity metric. For GED, we apply a one-to-one mapping function to transform the true distance score into the true similarity score; for MCS, the normalized MCS size is treated as the true similarity score.


\section{Experiments} 
\label{sec-exp}

\begin{table*}
\footnotesize
\centering
\caption{Effectiveness results on the GED metric. On \textsc{Aids} and \textsc{Linux}, \textsc{A*} provides ground-truth results, labeled with superscript $*$. On \textsc{Imdb} and \textsc{Ptc}, \textsc{A*} fails to compute most GEDs within a 5-minute limit (thus denoted as $-$). Instead, the minimum GED returned by \textsc{Beam}, \textsc{Hungarian}, and \textsc{VJ} for each pair is used as the ground-truth GED. The mse is in $10^{-3}$.
}
  \begin{tabular}{p{12mm}p{6mm}p{6mm}p{6mm}p{6mm}p{6mm}p{6mm}p{6mm}p{6mm}p{6mm}p{6mm}p{6mm}p{6mm}}
    \toprule
    \multirow{3}{*}{\textbf{Method}} &
      \multicolumn{3}{c}{\textbf{\textsc{Aids}}} &
      \multicolumn{3}{c}{\textbf{\textsc{Linux}}} &
      \multicolumn{3}{c}{\textbf{\textsc{Imdb}}} &
      \multicolumn{3}{c}{\textbf{\textsc{Ptc}}} \\
      & \textbf{mse} & \textbf{$\rho$} & \textbf{p@10} & \textbf{mse} & \textbf{$\rho$} & \textbf{p@10} & \textbf{mse} & \textbf{$\rho$} & \textbf{p@10} & \textbf{mse} & \textbf{$\rho$} & \textbf{p@10}\\
      \midrule
      
      
    \textsc{A*} & $0.000^{*}$ & $1.000^{*}$ & $1.000^{*}$ & $0.000^{*}$ & $1.000^{*}$ & $1.000^{*}$ & $-$ & $-$ & $-$ & $-$ & $-$ & $-$\\
    \textsc{Beam} & $12.090$ & $0.609$ & $0.481$ & $9.268$ & $0.827$ & $0.973$ & $2.413^{*}$ & $0.905^{*}$ & $0.803^{*}$ & $0.552^{*}$ & $0.998^{*}$ & $0.982^{*}$ \\
    \textsc{Hungarian} & $25.296$ & $0.510$ & $0.360$ & $29.805$ & $0.638$ & $0.913$ & $1.845^{*}$ & $0.932^{*}$ & $0.825^{*}$ & $112.326^{*}$ & $0.919^{*}$ & $0.159^{*}$ \\
    \textsc{VJ} & $29.157$ & $0.517$ & $0.310$ & $63.863$ & $0.581$ & $0.287$ & $1.831^{*}$ & $0.934^{*}$ & $0.815^{*}$ & $154.791^{*}$ &$0.904^{*}$ & $0.102^{*}$ \\
    \textsc{HED} & $28.925$ & $0.621$ & $0.386$ & $19.553$ & $0.897$ & $0.982$ & $19.400$ & $0.751$ & $0.801$ & $978.318$ & $0.919$ & $0.169$ \\ \hline
    \textsc{Smpnn} & $5.184$ & $0.294$ & $0.032$ & $11.737$ & $0.036$ & $0.009$ & $32.596$ & $0.107$ & $0.023$ & $116.473$ & $0.148$ & $0.082$ \\
    \textsc{EmbAvg} & $3.642$ & $0.601$ & $0.176$ & $18.274$ & $0.165$ & $0.071$ & $71.789$ & $0.229$ & $0.233$ & $32.601$ & $0.393$ & $0.173$ \\
    \textsc{GCNMean} & $3.352$ & $0.653$ & $0.186$ & $8.458$ &  $0.419$ & $0.141$ & $68.823$ & $0.402$ & $0.200$ & $6.525$ & $0.546$ & $0.150$ \\
    \textsc{GCNMax} & $3.602$ &  $0.628$ & $0.195$ & $6.403$ &  $0.633$ & $0.437$ & $50.878$ & $0.449$ & $0.425$ & $7.501$ & $0.506$ & $0.152$ \\
    \textsc{SimGNN} & $1.189$ & $0.843$ & $0.421$ & $1.509$ & $0.939$ & $0.942$ & $1.264$ & $0.878$ & $0.759$ & $0.850$ & $0.944$ & $0.507$ \\ 
    \textsc{GMN} & $1.886$ & $0.751$ & $0.401$ & $1.027$ & $0.933$ & $0.833$ & $4.422$ & $0.725$ & $0.604$ & $1.613$ & $0.672$ & $0.262$ \\ \hline
    \textsc{\model} & $\textbf{0.787}$ & $\textbf{0.874}$ & $\textbf{0.534}$ & $\textbf{0.058}$ & $\textbf{0.981}$ & $\textbf{0.992}$ & $\textbf{0.743}$ &  $\textbf{0.926}$ & $\textbf{0.828}$ & $\textbf{0.749}$ & $\textbf{0.956}$ & $\textbf{0.529}$ \\ \hline
  \end{tabular}
\centering
\label{table:ged_results}
\end{table*}

We evaluate our model, \model, against a number of state-of-the-art approaches to GED and MCS computation, with the major goals of addressing the following questions:
\begin{enumerate}
\item[{\bf Q1}] How accurate (effective) and fast (efficient) is \model compared to the state-of-the-art approaches for graph similarity computation, including both approximate similarity computation algorithms and neural network based models?
\item[{\bf Q2}] How do the proposed ordering, resizing, and multi-scale comparison techniques help with the CNN-based \model model?
\item[{\bf Q3}] Does \model yield meaningful and interpretable similarity matrices/images on the input graph pairs?
\end{enumerate}

\textbf{Datasets} \enspace To probe the ability of \model to compute graph-graph similarities from graphs in different domains, we evaluate on four real graph datasets, \textsc{Aids}, \textsc{Linux}, \textsc{Imdb}, and \textsc{Ptc}, whose detailed descriptions and statistics can be found in the supplementary material. 

For each dataset, we split it into training, validation, and testing sets by 6:2:2, and report the averaged \textit{Mean Squared Error (mse)}, \textit{Spearman's Rank Correlation Coefficient ($\rho$)}~\cite{spearman1904proof}, \textit{Kendall's Rank Correlation Coefficient ($\tau$)}~\cite{kendall1938new}, and \textit{Precision at $k$ (p@$k$)} to test the accuracy and ranking performance of each GED and MCS computation method. The supplementary material contains more details on the data preprocessing, parameter settings, result analysis, efficiency comparison, as well as parameter sensitivity study.

\textbf{Baseline Methods} \enspace We consider both the state-of-the-art GED/MCS computation methods and baselines using neural networks. To ensure consistency, all neural network models use GCN for node embeddings except for GMN, and to demonstrate the flexibility of our framework, we show the performance improvement of \model by replacing GCN with the more powerful GMN's node embedding methods in the supplementary material.

\subsection{Effectiveness}
\label{subsec-effectiveness}

\begin{table*}[h]
\footnotesize
\centering
\caption{Effectiveness results on the MCS metric. On all the datasets, \textsc{Mcsplit} provides ground-truth results, labeled with superscript $*$. The mse is in $10^{-3}$.}
  \begin{tabular}{p{12mm}p{6mm}p{6mm}p{6mm}p{6mm}p{6mm}p{6mm}p{6mm}p{6mm}p{6mm}p{6mm}p{6mm}p{6mm}}
    \toprule
    \multirow{3}{*}{\textbf{Method}} &
      \multicolumn{3}{c}{\textbf{\textsc{Aids}}} &
      \multicolumn{3}{c}{\textbf{\textsc{Linux}}} &
      \multicolumn{3}{c}{\textbf{\textsc{Imdb}}} &
      \multicolumn{3}{c}{\textbf{\textsc{Ptc}}} \\
      & \textbf{mse} & \textbf{$\rho$} & \textbf{p@10} & \textbf{mse} & \textbf{$\rho$} & \textbf{p@10} & \textbf{mse} & \textbf{$\rho$} & \textbf{p@10} & \textbf{mse} & \textbf{$\rho$} & \textbf{p@10}\\
      \midrule
      
      
    \textsc{Mcsplit} & $0.000^{*}$ & $1.000^{*}$ & $1.000^{*}$ & $0.000^{*}$ & $1.000^{*}$ & $1.000^{*}$ & $0.000^{*}$ & $1.000^{*}$ & $1.000^{*}$ & $0.000^{*}$ & $1.000^{*}$ & $1.000^{*}$ \\ \hline
    \textsc{Smpnn} & $4.592$ & $0.755$ & $0.348$ & $3.558$ & $0.126$ & $0.236$ & $11.018$ & $0.330$ & $0.003$ & $11.001$ & $0.502$ & $0.146$ \\
    \textsc{EmbAvg} & $6.466$ & $0.701$ & $0.236$ & $2.663$ & $0.427$ & $0.343$ & $17.853$ & $0.524$ & $0.166$ & $23.018$ & $0.556$ & $0.302$ \\
    \textsc{GCNMean} & $5.956$ & $0.776$ & $0.316$ & $2.706$ & $0.439$ & $0.368$ & $9.316$ & $0.753$ & $0.364$ & $9.166$ & $0.712$ & $0.337$ \\
    \textsc{GCNMax} & $5.525$ &  $0.782$ & $0.328$ & $2.466$ & $0.543$ & $0.440$ & $8.234$ & $0.796$ & $0.435$ & $7.905$ & $0.752$ & $0.385$ \\
    \textsc{SimGNN} & $3.433$ & $0.822$ & $0.374$ & $0.729$ & $0.859$ & $0.850$ & $1.153$ & $0.938$ & $0.705$ & $3.781$ & $0.782$ & $0.279$ \\
    \textsc{GMN} & $\textbf{1.750}$ & $\textbf{0.909}$ & $\textbf{0.591}$ & $0.598$ & $0.906$ & $0.830$ & $0.590$ & $0.941$ & $0.795$ &  $3.835$ & $0.841$ & $\textbf{0.530}$ \\ \hline
    \textsc{\model} & $2.402$ & $0.858$ & $0.505$ & $\textbf{0.164}$ & $\textbf{0.962}$ & $\textbf{0.951}$ & $\textbf{0.307}$ & $\textbf{0.976}$ & $\textbf{0.817}$ & $\textbf{2.268}$ & $\textbf{0.854}$ & $0.504$ \\ \hline
  \end{tabular}
\centering
\label{table:mcs_results}
\end{table*}

As shown in Table~\ref{table:ged_results} and \ref{table:mcs_results}, our model, \model, consistently achieves the best results on all metrics across all the datasets with both the GED and MCS metrics. Specifically, \model achieves the smallest error and the best ranking performance on the task of graph similarity computation under six out of eight settings. We repeated the running of our model 10 times on \textsc{Aids}, and the standard deviation of mse is $4.56*10^{-5}$. The \textsc{Aids} dataset is relatively small in terms of the average number of nodes per graph, potentially causing the CNN model to overfit. In the supplementary material, we show that as a general framework, by replacing GCN with \textsc{GMN}'s node embedding method, \model achieves the best performance than all methods.

\subsection{Efficiency}
\label{subsec-efficiency}
\begin{figure}[h]
\centering
\includegraphics[width=.95\columnwidth]{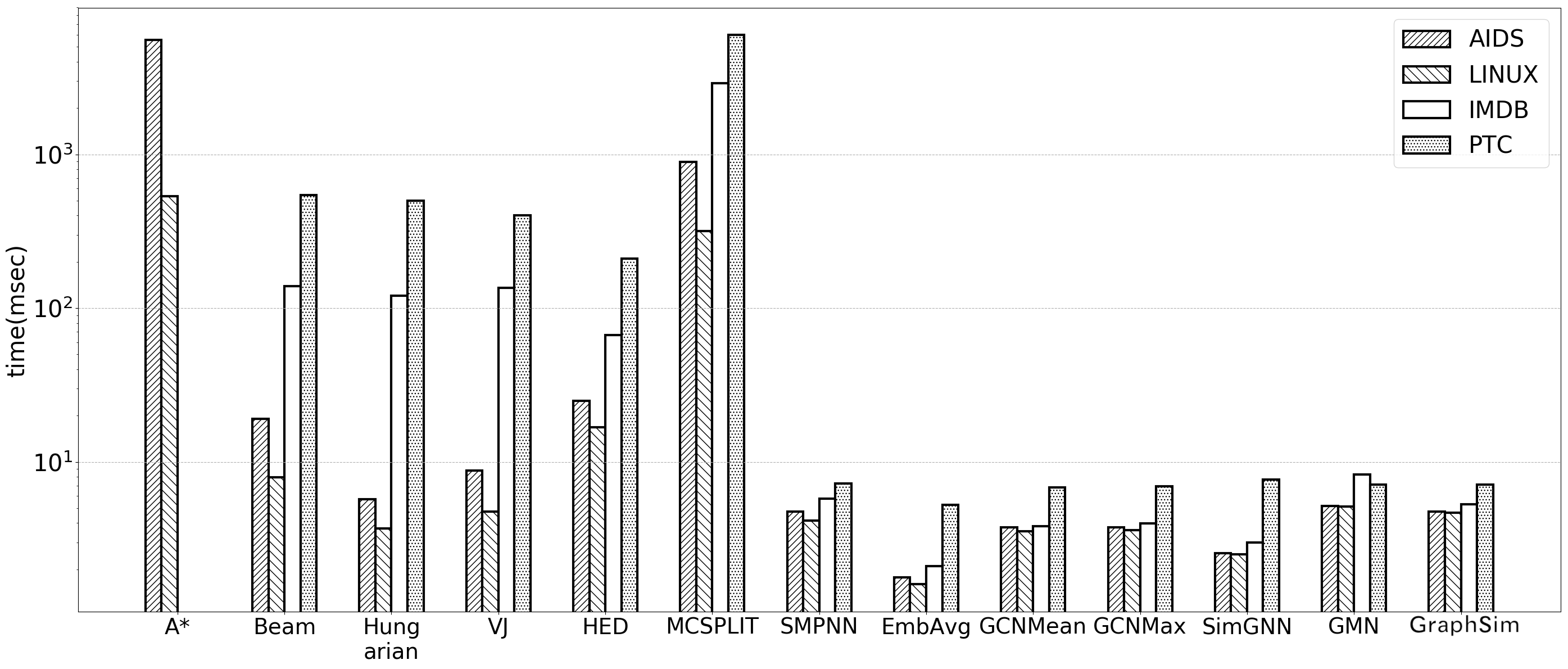}
\caption{Running time comparison. The y-axis uses the log scale. The running time is averaged across queries. For neural network models, the running time for GED and MCS computation is very close to each other, so we take the average of the two.}
\label{fig:time}
\end{figure}

In Fig.~\ref{fig:time}, the results are averaged across queries and in wall time. \textsc{EmbAvg} is the fastest method among all, but its performance is poor, since it simply takes the dot product between two graph-level embeddings (average of node embeddings) as the predicted similarity score. \textsc{Beam} and \textsc{Hungarian} run fast on \textsc{Linux}, but due to their higher time complexity, they scale poorly on the largest dataset, \textsc{Imdb}. The exact MCS solver \textsc{Mcsplit} is the state-of-the-art for MCS computation, and runs faster than the exact GED solver, \textsc{A*} on all datasets. However, in general, neural network based models are still much faster than these solvers, since they enjoy lower time complexity in general and additional benefits from parallelizability and acceleration provided by GPU.

\subsection{Analysis of Various Proposed Techniques in \textsc{\model}}

To address {\bf Q2}, we conduct several experiments by comparing \model with three simpler variants, whose results are shown in Table~\ref{table:extra_results}. Among the three proposed techniques (i.e., max padding plus matrix resizing, ordering, and multi-scale comparison), max padding and matrix resizing affects the performance the most: Comparing \textsc{GS-Pad} with \textsc{GS-Resize} on \textsc{Imdb} and \textsc{Ptc}, our proposed padding and resizing technique greatly reduces the approximation error. Such significant improvements on \textsc{Imdb} and \textsc{Ptc} can be attributed to the large average graph size and graph size variance, as seen from the dataset statistics in the supplementary material. Besides, by comparing the \textsc{GS-Resize} and \textsc{\model}, we can see a performance boost brought by the multi-scale framework. The advantage of BFS ordering can be observed by comparison of \textsc{GS-NoOrd} and \textsc{\model}.


\begin{table*}
\footnotesize
\centering

\caption{\textsc{GS-Pad} and \textsc{GS-Resize} perform node ordering and use the node embeddings only by the last GCN layer to generate the similarity matrix. Since CNNs require fixed-length input, if the model does not use resizing, a simple way is to zero pad each similarity matrix to $N_{\mathrm{max}}$ by $N_{\mathrm{max}}$ (maximum graph size in the entire dataset), denoted as \textsc{GS-Pad}. \textsc{GS-Resize} uses the proposed max padding and resizing techniques. \textsc{GS-NoOrd} uses all the proposed techniques including multi-scale comparison except for node ordering. The results are on the GED metric. The mse is in $10^{-3}$.}
  \begin{tabular}{p{16.5mm}p{6mm}p{6mm}p{6mm}p{6mm}p{6mm}p{6mm}p{6mm}p{6mm}p{6mm}p{6mm}p{6mm}p{6mm}}
    \toprule
    \multirow{3}{*}{\textbf{Method}} &
      \multicolumn{3}{c}{\textbf{\textsc{Aids}}} &
      \multicolumn{3}{c}{\textbf{\textsc{Linux}}} &
      \multicolumn{3}{c}{\textbf{\textsc{Imdb}}} &
      \multicolumn{3}{c}{\textbf{\textsc{Ptc}}} \\
      & \textbf{mse} & \textbf{$\rho$} & \textbf{p@10} & \textbf{mse} & \textbf{$\rho$} & \textbf{p@10} & \textbf{mse} & \textbf{$\rho$} & \textbf{p@10} & \textbf{mse} & \textbf{$\rho$} & \textbf{p@10}\\
      \midrule
      
    \textsc{GS-Pad} & 0.807 & 0.863 & 0.514 & 0.141 & 0.988 & 0.983 & 6.455 & 0.661 & 0.552 & 6.808 & 0.637 & 0.481 \\
    \textsc{GS-Resize} & 0.811 & 0.866 & 0.499 & 0.103 & 0.991 & 0.986 & 0.896 & 0.914 & 0.810 & 0.791 & 0.948 & 0.513 \\
    \textsc{GS-NoOrd} & 0.803 & 0.867 & 0.478 & 0.071 & 0.983 & 0.976 & 1.105 & 0.902 & 0.744 & 0.838 & 0.945 & 0.515 \\
    \midrule
    \model & \textbf{0.787} &
    \textbf{0.874} &
    \textbf{0.534} &
    \textbf{0.058} &
    \textbf{0.993} &
    \textbf{0.981} &
    \textbf{0.743} &
    \textbf{0.926} &
    \textbf{0.828} &
    \textbf{0.749} &
    \textbf{0.956} &
    \textbf{0.529}
 \\ \hline
  \end{tabular}
\centering
\label{table:extra_results}
\end{table*}

\subsection{Analysis of Similarity ``Images'' in \textsc{\model}}

\begin{figure}
\centering
\includegraphics[width=.95\columnwidth]{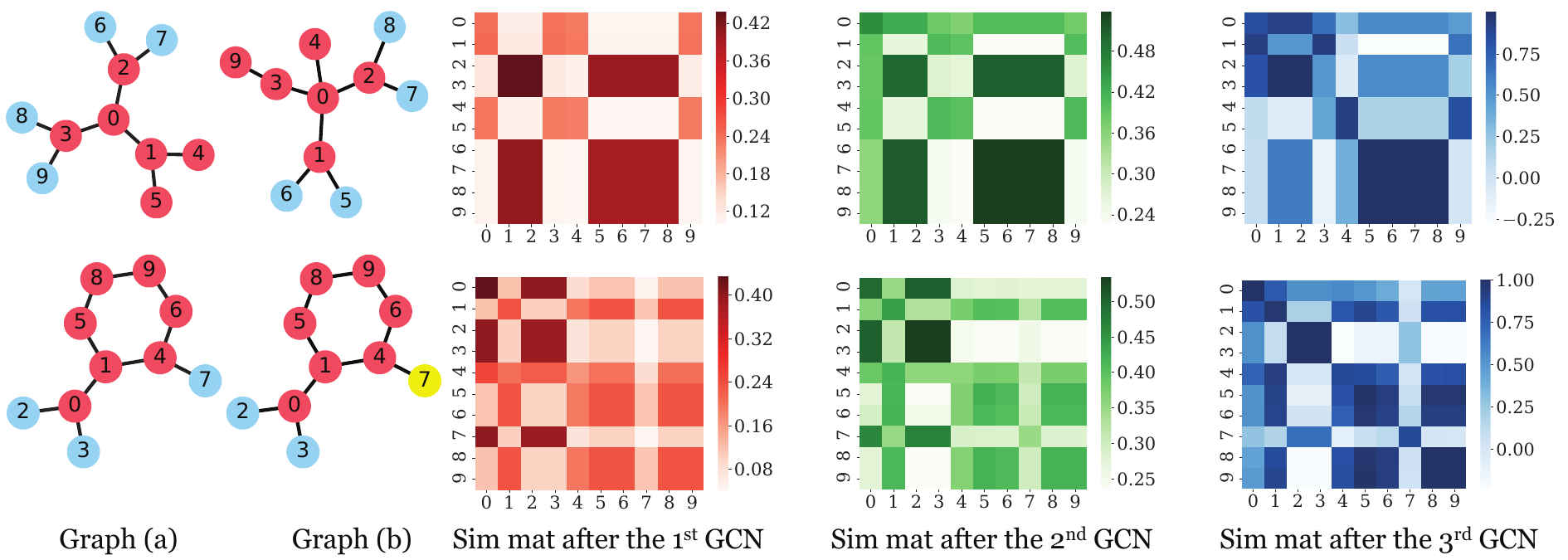}
\caption{Similarity ``images'' on two pairs of graphs from \textsc{Aids} trained with the GED metric. Different heatmap colors differentiate ``images'' from different scales of comparison.
Top pair: GED=2---an edge deletion (edge between node 0 and 4 of graph (b)) and an edge addition (edge between node 3 and 4 of graph (b)). Bottom pair: GED=1---a node relabeling (node 7).
}
\label{fig:multi_scale_heatmaps}
\end{figure}


We demonstrate six similarity matrices (plotted as heatmap images) generated by \model on \textsc{Aids} in Fig.~\ref{fig:multi_scale_heatmaps}. Node ids come from BFS ordering. Since both pairs are quite similar, as mentioned in Section~\ref{subsec-cnn}, we expect to see block patterns, which are actually observed in the six similarity matrices. 
From pair (1) (the top row), we can see that larger scale (the blue matrix) helps distinguish between nodes 3 and 4 of graph (b), which contributes to the GED edit sequence as shown in the caption of Fig~\ref{fig:multi_scale_heatmaps}. From pair (2) (the bottom row), we can see that the smaller scale (the red matrix) helps differentiate between node 7 in graph (a) and node 7 in graph (b). Notice that comparing at larger scales does not help tell their difference in node types, because their structural equivalence causes their similarities to be higher as seen in the green and blue matrices. Thus, Fig.~\ref{fig:multi_scale_heatmaps} shows the importance of using multi-scale similarity matrices rather than a single one. 


\subsection{Case Studies}

\begin{figure}[h!]
    \centering
    
    {{\includegraphics[width=.82\columnwidth]{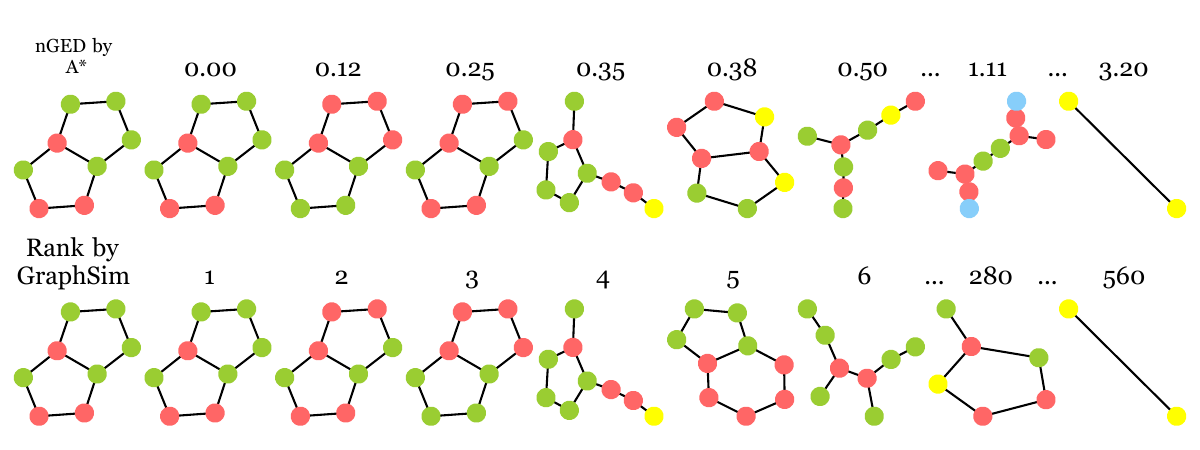} }}
    \centering
    {{\includegraphics[width=.82\columnwidth]{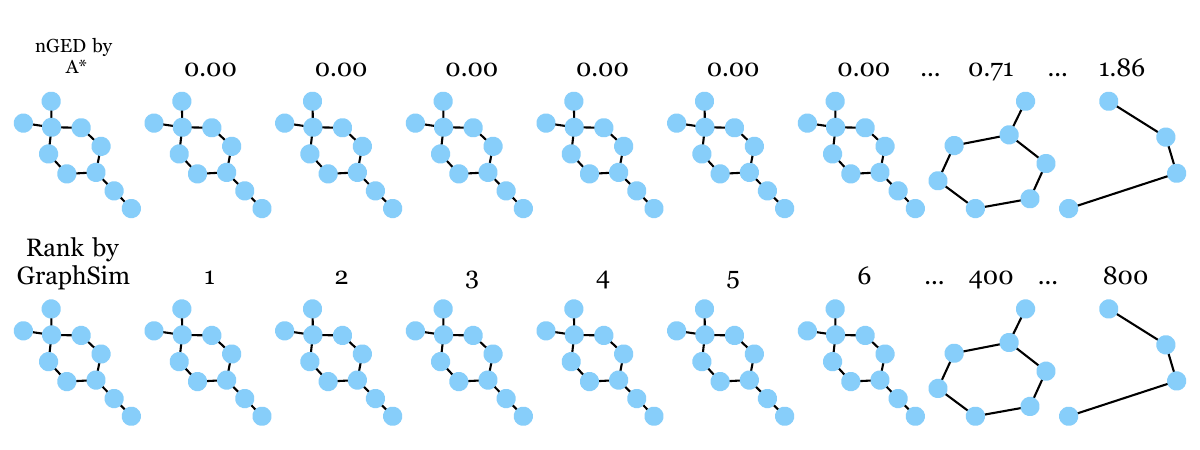} 
    \label{fig:search_demo_Linux}}}
    \centering
    {{\includegraphics[width=.82\columnwidth]{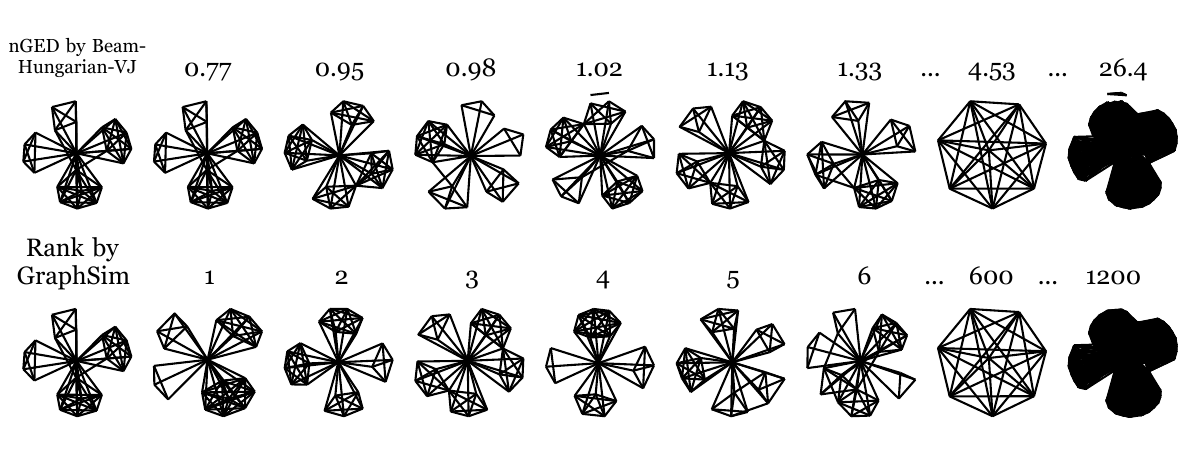} }}
    {{\includegraphics[width=.82\columnwidth]{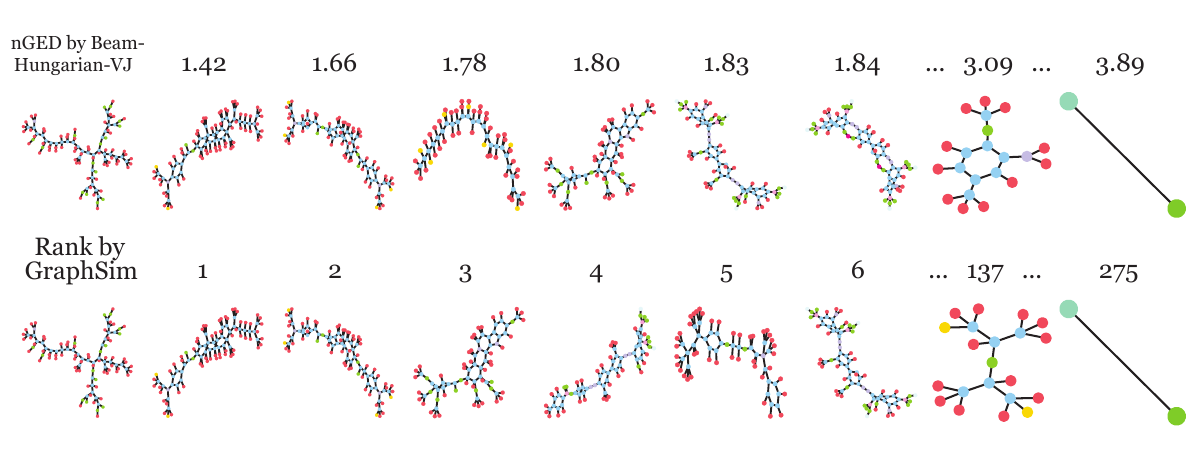} 
    \label{fig:search_demo_Ptc}}}
    
    \centering
    \caption{Visualization of ranking results under the GED metric. From top to bottom: \textsc{Aids}, \textsc{Linux}, \textsc{Imdb}, \textsc{Ptc}.}
    \label{fig:search_demo}
\end{figure}

We demonstrate four example queries, one from each dataset in Fig.~\ref{fig:search_demo}. In each demo, the top row depicts the query along with the ground-truth ranking results, labeled with their normalized GEDs to the query. The bottom row shows the graphs returned by our model, each with its rank shown at the top. \model is able to retrieve graphs similar to the query. For example, in the case of LINUX, the top 6 results are exactly the isomorphic graphs to the query.
\section{Conclusion}
\label{sec-conc}
We introduced a CNN based method for better graph similarity computation. Using \model in conjunction with existing methods for generating node embeddings, we improve the performance in several datasets on two graph proximity metrics: Graph Edit Distance (GED) and Maximum Common Subgraph (MCS). Interesting future directions include using hierarchical graph representation learning techniques to reduce computational time complexity involved in the node-node similarity computation, and applying the CNN based graph matching method to other graph matching tasks, e.g. network alignment, as well as the explicit generation of edit sequence for GED and node correspondence for MCS.

\subsubsection*{Acknowledgments}

This work is partially supported by NSF III-1705169, NSF CAREER Award 1741634, NSF \#1937599, Okawa Foundation Grant, and Amazon Research Award.

\bibliography{bibliography}
\bibliographystyle{aaai}

\setcounter{section}{0} 
\clearpage
\newpage
\section*{{\Large Supplementary Material}}

\def\thesection{\Alph{section}}

\graphicspath{ {images/} }

\section{Convolutional Set Matching}
\label{sec-theory}

In this section, we present \model from the perspective of set matching, by making theoretical connections with two types of graph matching methods: optimal assignment kernels for graph classification and bipartite graph matching for GED computation. In fact, beyond graphs, set matching has broader applications in Computer Networking (e.g. Internet content delivery)~\cite{maggs2015algorithmic}, Computer Vision (e.g. semantic visual matching)~\cite{zanfir2018deep}, Bioinformatics (e.g. protein alignment)~\cite{zaslavskiy2009global}, Internet Advertising (e.g. advertisement auctions)~\cite{edelman2005advertising}, Labor Markets (e.g. intern host matching)~\cite{roth1984medical}, etc. This opens massive possibilities for future work and suggests the potential impact of \model beyond the graph learning community. 

\subsection{Connection with Optimal Assignment Kernels}

Graph kernels measure the similarity between two graphs, and have been extensively applied to the task of graph classification. 
Among different families of graph kernels, optimal assignment kernels establish the correspondence between parts of the two graphs, and have many variants~\cite{frohlich2005optimal,johansson2015learning,kriege2016valid,nikolentzos2017matching}. 
Intuitively, the optimal assignment graph kernels maximize the total similarity between the assigned parts. 

Let us take the Earth Mover's Distance (\textsc{EMD}) kernel~\cite{nikolentzos2017matching} as an example, since it is among the most similar methods to our proposed approach. It treats a graph as a bag of node embedding vectors (not trainable), and computes the optimal ``travel cost'' between two graphs, where the cost is defined as the $L$-2 distance between node embeddings. Given two graphs with node embeddings $\bm{X} \in \mathbb{R}^{N_1 \times D}$ and $\bm{Y} \in \mathbb{R}^{N_2 \times D}$, it solves the following transportation problem~\cite{rubner2000earth}:
\begin{equation}
\label{eq:mne-theory}
\begin{aligned}
\min & \sum_{i=1}^{N_1} \sum_{j=1}^{N_2} \bm{T}_{ij} {||\bm{x}_i - \bm{y}_j||}_2 \\
& \mathrm{subject \ to} \\
& \sum_{i=1}^{N_1} \bm{T}_{ij} = \frac{1}{N_2} \quad \forall j \in \{ 1,...,N_2 \} \\
& \sum_{j=1}^{N_2} \bm{T}_{ij} = \frac{1}{N_1} \quad \forall i \in \{ 1,...,N_1 \} \\
& \bm{T}_{ij} \geq 0 \quad \forall i \in \{ 1,...,N_1 \}, \forall j \in \{ 1,..,N_2 \}
\end{aligned}
\end{equation}

subject to constraints that each row and column of $\bm{T}$ must sum up to $\frac{1}{N_2}$ and $\frac{1}{N_1}$, respectively, and each entry of $\bm{T}$ must be non-negative. $\bm{T} \in \mathbb{R}^{N_1 \times N_2}$ denotes the flow matrix, with $\bm{T}_{ij}$ being how much of node $i$ in $\mathcal{G}_1$ travels (or ``flows'') to node $j$ in $\mathcal{G}_2$. In other words, the \textsc{EMD} between two graphs is the minimum amount of “work” that needs to be done to transform one graph to another, where the optimal transportation plan is encoded by $\bm{T}^{*}$. 

It has been shown that if $N_1=N_2=N$, the optimal solution satisfies $\bm{T}_{ij}^{*} \in \{0,\frac{1}{N}\}$~\cite{balinski1961fixed}, satisfying the optimal bijection requirement of the assignment kernel. Even if $N_1 \neq N_2$, this can still be regarded as approximating an assignment problem~\cite{fan2017point}.

To show the relation between the \textsc{EMD} kernel and our approach, we consider \model as a mapping function that, given two graphs with node embeddings $\bm{X} \in \mathbb{R}^{N_1 \times D}$ and $\bm{Y} \in \mathbb{R}^{N_2 \times D}$, produces one score as the predicted similarity score, which is compared against the true similarity score:
\begin{equation}
\label{eq:mne_equivalent-theory}
\begin{aligned}
\min (h_{\bm{\Theta}} (\bm{X}, \bm{Y}) - s_{12})^2
\end{aligned}
\end{equation}
where $h_{\bm{\Theta}} (\bm{X}, \bm{Y})$ represents the similarity matrix generation layer and CNN layers but can potentially be replaced by any neural network transformation.

To further illustrate the connection, we consider one CNN layer with one filter of size $N$ by $N$, where $N = \mathrm{max}(N_1, N_2)$. Then Eq.~\ref{eq:mne_equivalent-theory} becomes:
\begin{equation}
\begin{aligned}
\min (\sigma (\sum_{i=1}^{N} \sum_{j=1}^{N} \bm{\Theta}_{ij} (\bm{x}_i^{T} \bm{y}_j)) - s_{12})^2
\end{aligned}
\end{equation}
where $\bm{\Theta} \in \mathbb{R}^{N \times N}$ is the convolutional filter.

Compared with the \textsc{EMD} kernel, our method has two benefits. (1) The mapping function and the node embeddings $\bm{X}$ and $\bm{Y}$ are simultaneously learned through backpropagation, while the kernel method solves the assignment problem to obtain $\bm{T}$ and uses un-trainable node embeddings 
$\bm{X}$ and $\bm{Y}$, e.g. generated by the decomposition of the graph Laplacian matrix. Thus, \model is suitable for \textit{learning} an approximation of a graph distance metric, while the kernel method cannot. The typical usage of a graph kernel is to feed the graph-graph similarities into a SVM classifier for graph classification. (2) The best average time complexity of solving Eq.~\ref{eq:mne-theory} scales $O(N^3 log N)$~\cite{pele2009fast}, where $N$ denotes the
number of total nodes in two graphs, while the convolution operation is in O($(\mathrm{max}(N_1,N_2))^2$) time.

\subsection{Connection with Bipartite Graph Matching}
\label{subsec-bgm}

Among the existing approximate GED computation algorithms, Hungarian~\cite{riesen2009approximate} and VJ~\cite{fankhauser2011speeding} are two classic ones based on bipartite graph matching. Similar to the optimal assignment kernels, Hungarian and VJ also find an optimal match between the nodes of two graphs. However, different from the \textsc{EMD} kernel, the assignment problem has stricter constraints: One node in $\mathcal{G}_1$ can be only mapped to one other node in $\mathcal{G}_2$. Thus, the entries in the assignment matrix $\bm{T} \in \mathbb{R}^{N^{'} \times N^{'}}$ are either 0 or 1, denoting the operations transforming $\mathcal{G}_1$ into $\mathcal{G}_2$, where $N^{'} = N_1 + N_2$. The assignment problem takes the following form:
\begin{equation}
\label{eq:bgm-theory}
\begin{aligned}
\min & \sum_{i=1}^{N^{'}} \sum_{j=1}^{N^{'}} \bm{T}_{ij} \bm{C}_{ij} \\
& \mathrm{subject \ to} \\
& \sum_{i=1}^{N^{'}} \bm{T}_{ij} = 1 \quad \forall j \in \{ 1,...,N^{'} \} \\
& \sum_{j=1}^{N^{'}} \bm{T}_{ij} = 1 \quad \forall i \in \{ 1,...,N^{'} \} \\
& \bm{T}_{ij} \in \{0, 1\} \quad \forall i \in \{ 1,...,N^{'} \}, \forall j \in \{ 1,..,N^{'} \}
\end{aligned}
\end{equation}
subject to constraints that each row and column of $\bm{T}$ must sum up to 1 and each entry of $\bm{T}$ must be either 0 or 1. The cost matrix $\bm{C} \in \mathbb{R}^{N^{'} \times N^{'}}$ reflects the GED model and is defined as follows:
\[
\renewcommand\arraystretch{1.3}
\bm{C} = \mleft[
\begin{array}{ccc|ccc}
  \bm{C}_{1,1} & \dots & \bm{C}_{1,N_2} & \bm{C}_{1,\epsilon} & \dots & \infty \\
  \vdots & \ddots & \vdots & \vdots & \ddots & \vdots \\
  \bm{C}_{N_1,1} & \dots & \bm{C}_{N_1,N_2} & \infty & \dots & \bm{C}_{N_1,\epsilon} \\
  \hline
  \bm{C}_{\epsilon,1} & \dots & \infty & 0 & \dots & 0 \\
  \vdots & \ddots & \vdots & \vdots & \ddots & \vdots \\
  \infty & \dots & \bm{C}_{\epsilon,N_2} & 0 & \dots & 0 \\
\end{array}
\mright]
\]
where $\bm{C}_{i,j}$ denotes the cost of a substitution, $\bm{C}_{i,\epsilon}$ denotes the cost of a node deletion, and $\bm{C}_{\epsilon,j}$ denotes the cost of a node insertion. According to our GED definition, $\bm{C}_{ij} = 0$ if the labels of node $i$ and node $j$ are the same, and 1 otherwise; $\bm{C}_{i,\epsilon} = \bm{C}_{\epsilon,j} = 1$.

Exactly solving this constrained optimization program would yield the exact GED solution~\cite{fankhauser2011speeding}, but it is NP-complete since it is equivalent to finding an optimal matching in a complete bipartite graph~\cite{riesen2009approximate}. 


To efficiently solve the assignment problem, the \textsc{Hungarian} algorithm~\cite{kuhn1955hungarian} and \textsc{Volgenant Jonker (VJ)}~\cite{jonker1987shortest} algorithm are commonly used, and both both run in cubic time. In contrast, \model takes advantage of the exact solutions of this problem during the training stage, and computes the approximate GED 
in quadratic time, without the need for solving any optimization problem for a new graph pair.

\subsection{Set Matching Based Graph Similarity Computation}
\label{subsec:set-matching-gsc}




In fact, the forward pass of \model can be interpreted as a two-step procedure: 1. Applying a GCN-based graph transforming function ($f(\cdot)$) to transform each graph into multiple sets of node embeddings; 2. Applying multiple set matching functions $g(\cdot,\cdot)$s to compare the node embeddings at different scales. Thus, we call our approach set matching based graph similarity computation, or convolutional set matching as in Table~\ref{table:set_matching_summary}.

In contrast, the other two approaches in Table~\ref{table:set_matching_summary} also model the graph-graph similarity by viewing a graph as a set, but suffer from limited learnability and cannot be trained end-to-end. Due to its neural network nature, the convolutional set matching approach enjoys flexibility and thus has the potential to be extended to solve other set matching problems.

\begin{table*}[h]
\footnotesize
\centering
\caption{Summary of three set matching based approaches to graph similarity. $f(\cdot)$ denotes the graph transforming function, and $g(\cdot,\cdot)$ denotes the set matching function. 
}
\begin{tabular}
{p{45mm}p{45mm}p{25mm}p{25mm}} \hline
\textbf{Approach} & \textbf{Example(s)} & \textbf{$f(\mathcal{G})$} & \textbf{$g(\mathcal{G}_1,\mathcal{G}_2)$} \\
\hline
\textbf{Optimal Alignment Kernels} & \textsc{EMD} kernel~\cite{nikolentzos2017matching} & Node Embedding & Solver of Eq.~\ref{eq:mne} \\
\textbf{Bipartite Graph Matching} & \textsc{Hungarian}~\cite{riesen2009approximate}, \textsc{VJ}~\cite{fankhauser2011speeding} & Nodes of $\mathcal{G}$ & Solver of Eq.~\ref{eq:bgm-theory} \\
\textbf{Convolutional Set Matching} & \model & Node Embedding & Sim. Matrix + CNN \\ \hline
\end{tabular}
\centering
\label{table:set_matching_summary}
\end{table*}

\section{Datasets}
Four real-world graph datasets are used in the experiments. A concise summary can be found in Table~\ref{table:dataset_summary}.

\begin{table*}[h]
\footnotesize
\centering
\caption{Statistics of datasets. ``Min'', ``Max'', ``Mean'', and ``Std'' refer to the minimum, maximum, mean, and standard deviation of the graph sizes (number of nodes), respectively.}
\begin{tabular}
{lccccccc} \hline
\textbf{Dataset} & \textbf{Meaning} & \textbf{\#Node Labels} & \textbf{\#Graphs} & \textbf{Min} & \textbf{Max} & \textbf{Mean} & \textbf{Std}\\ \hline
\textbf{\textsc{Aids}} & Chemical Compounds & 29 & 700 & 2 & 10 & 8.9 & 1.4 \\
\textbf{\textsc{Linux}} & Program Dependence Graphs & 1 & 1000 & 4 & 10 & 7.7 & 1.5  \\
\textbf{\textsc{Imdb}} & Ego-Networks & 1 & 1500 & 7 & 89 & 13.0 & 8.5 \\
\textbf{\textsc{Ptc}} & Chemical Compounds & 19 & 344 & 2 & 109 & 25.6 & 16.2 \\
\hline
\end{tabular}
\centering
\label{table:dataset_summary}
\end{table*}

\begin{itemize}
    \item \textsc{Aids}~\cite{zeng2009comparing,wang2012efficient} consists of 42,687 chemical compounds from the Developmental Therapeutics Program at NCI/NIH 7~\footnote{\url{https://wiki.nci.nih.gov/display/NCIDTPdata/Aids+Antiviral+Screen+Data}}, out of which we select 700 graphs, where each node is labeled with one of 29 types. The Hydrogen atoms of graph compound is omitted in the graph representation.
    
    \item \textsc{Linux}~\cite{wang2012efficient} consists of 48,747 Program Dependence Graphs (PDG) generated from the \textsc{Linux} kernel, out of which we select 1000 graphs. Each graph represents a function, where a node represents one statement and an edge represents the dependency between the two statements. The nodes are unlabelled.
    
    \item \textsc{Imdb}~\cite{yanardag2015deep} consists of 1500 ego-networks of movie actors/actresses with unlabeled nodes representing the people and edges representing the collaboration relationship. The nodes are unlabeled. 
    
     \item \textsc{Ptc}~\cite{shrivastava2014new} is also a collection of 344 chemical compounds similar to \textsc{Aids}. Although it has smaller number of graphs compared to \textsc{Aids}, it has larger graphs on average.

\end{itemize}

Since the GED/MCS computation is pairwise, it is necessary to take the number of pairs into consideration. There are 490K, 1M, 2.25M, and 118K graph pairs in the \textsc{Aids}, \textsc{Linux}, \textsc{Imdb} and \textsc{Ptc} datasets, respectively. Note that the four datasets are from three different domains.



\section{Data Preprocessing}
\label{subsec-data-preproc}

For each dataset, we randomly split 60\%, 20\%, and 20\% of all the graphs as training set, validation set, and testing set, respectively. For each graph in the testing set, we treat it as a query graph, and let the model compute the similarity between the query graph and every graph in the training and validation sets. 

Since graphs from \textsc{Aids} and \textsc{Linux} are relatively small, the exponential-time exact GED computation algorithm, \textsc{A*}~\cite{hart1968formal}, is used to compute the GEDs between all graph pairs. For the \textsc{Imdb} dataset, however, \textsc{A*} can no longer be used, as no currently available algorithm can reliably compute GED between graphs with more than 16 nodes within a reasonable time ~\cite{blumenthal2018exact}. Instead, we take the minimum distance computed by \textsc{Beam}~\cite{neuhaus2006fast}, \textsc{Hungarian}~\cite{riesen2009approximate}, and \textsc{VJ}~\cite{fankhauser2011speeding}. The minimum is taken because their returned GEDs are guaranteed to be upper bounds of the true GEDs. In fact, the ICPR 2016 Graph Distance Contest~\footnote{\url{https://gdc2016.greyc.fr/}} also adopts this approach to handle large graphs.

To transform true GEDs into true similarity scores to train our model, we first normalize the GEDs: $\mathrm{nGED}(\mathcal{G}_1,\mathcal{G}_2)=\frac{\mathrm{GED}(\mathcal{G}_1,\mathcal{G}_2)} {(|\mathcal{G}_1| + |\mathcal{G}_2|) / 2}$, where $|\mathcal{G}_i|$ denotes the number of nodes of $\mathcal{G}_i$~\cite{qureshi2007graph}, then adopt the exponential function $\lambda(x) = e^{-x}$, an one-to-one function, to transform the normalized GED into a similarity score in the range of $(0, 1]$.

We use \textsc{Mcsplit}~\cite{mccreesh2017partitioning}, a recent exact MCS solver, to generate ground-truth for the MCS metric. Since MCS is already a graph similarity metric, we simply normalize the MCS size and treat the normalized MCS size as the true similarity score: $\mathrm{nMCS}(\mathcal{G}_1,\mathcal{G}_2)=\frac{|\mathrm{MCS}(\mathcal{G}_1,\mathcal{G}_2)|} {(|\mathcal{G}_1| + |\mathcal{G}_2|) / 2}$. 

Please note that currently we are unaware of any approximate MCS algorithms that satisfy the exact MCS definition as used by the most state-of-the-art solver, \textsc{Mcsplit}. However, given the great efficiency improvement as shown in the main text
and the good accuracy as analyzed in the main text, \model demonstrates promising applicability in real world tasks where fast and accurate approximation is needed.

\section{Model Configurations} 
\label{subsec-param-set}

For the proposed model, to make a fair comparison with baselines, we use a single network architecture on all the datasets, and run the model using exactly the same test graphs as used in the baselines. We set the number of GCN layers to 3, and use ReLU as the activation function. The output dimensions for the 1st, 2nd, and 3rd layers of GCN are 128, 64, and 32, respectively. Notice that the only exception to this is \textsc{GMN}, which uses a cross-graph attention mechanism to generate its node embeddings. 

For the padding scheme, graphs in \textsc{Aids}, \textsc{Linux}, \textsc{Imdb}, and \textsc{Ptc} are padded to 10, 10, 90, and 109 nodes, respectively. For the resizing scheme, the similarity matrices for \textsc{Aids}, \textsc{Linux}, \textsc{Imdb}, and \textsc{Ptc} are resized to 10 by 10, 10 by 10, 54 by 54, and 54 by 54, respectively. 

For the CNNs, we use the following architecture: conv(6,1,1,16), maxpool(2), conv(6,1,16,32), maxpool(2), conv(5,1,32,64), maxpool(2), conv(5,1,64,128), maxpool(3), conv(5,1,128,128), maxpool(3) (``conv( window size, kernel stride, input channels, output channels )''; ``maxpool( pooling size )'').

We conduct all experiments on a single machine with an Intel i7-6800K CPU and one Nvidia Titan GPU. As for training, we set the batch size to 128, use the Adam algorithm for optimization~\cite{kingma2014adam}, and fix the initial learning rate to 0.001. We set the number of iterations to 15000, and select the best model based on the lowest validation loss.

\section{Effectiveness Comparison with Baselines}

\begin{table*}[h]
\footnotesize
\centering
\caption{\model combined with \textsc{GMN}'s node embedding method aceheievs the best performance on \textsc{Aids} under both the GED and MCS metrics. The mse is in $10^{-3}$.}
  \begin{tabular}{lllllll}
    \toprule
    \multirow{3}{*}{\textbf{Method}} &
      \multicolumn{3}{c}{\textbf{\textsc{Aids} (GED)}} &
      \multicolumn{3}{c}{\textbf{\textsc{Aids} (MCS)}} \\
      & \textbf{mse} & \textbf{$\rho$} & \textbf{p@10} & \textbf{mse} & \textbf{$\rho$} & \textbf{p@10} \\
      \midrule
      
    \textsc{GMN} & $1.886$ & $0.751$ & $0.401$ & $1.750$ & $0.909$ & $0.591$ \\
    \textsc{\model} & $0.787$ & $0.874$ & $0.534$ & $2.402$ & $0.858$ & $0.505$ \\
    \textsc{\model-w/\textsc{GMN}} & $\textbf{0.557}$ & $\textbf{0.907}$ & $\textbf{0.653}$ & $\textbf{1.244}$ & $\textbf{0.923}$ & $\textbf{0.615}$ \\ \hline
  \end{tabular}
\centering
\label{table:aids_with_gmn}
\end{table*}

\begin{table*}[h]
\footnotesize
\centering
\caption{Summary of time complexity of all methods used in the experiments. ``NN'' denotes whether the method is based on neural networks.}
\begin{tabular}
    {lccccc} \hline
    \textbf{Method} & \textbf{GED} & \textbf{MCS} & \textbf{Exact} & \textbf{NN} & \textbf{Worst-Case Time} \\ \hline
    \textsc{A*}~\cite{hart1968formal} & $\checkmark$ & $\times$ & $\checkmark$ & $\times$ & {$O({N_1}^{N_2})$} \\
    \textsc{Beam}~\cite{neuhaus2006fast} & $\checkmark$ & $\times$ & $\times$ & $\times$ & {$O({N_1}^{N_2})$} \\
    \textsc{Hungarian}~\cite{riesen2009approximate} & $\checkmark$ & $\times$ & $\times$ & $\times$ & {$O(({N_1}+{N_2})^3)$} \\ 
    \textsc{VJ}~\cite{fankhauser2011speeding} & $\checkmark$ & $\times$ & $\times$ & $\times$ & {$O(({N_1}+{N_2})^3)$} \\
    \textsc{HED}~\cite{fischer2015approximation} & $\checkmark$ & $\times$ & $\times$ & $\times$ & {$O(({N_1}+{N_2})^2)$} \\
    \textsc{Mcsplit}~\cite{mccreesh2017partitioning} & $\times$ & $\checkmark$ & $\checkmark$ & $\times$ & {$O({N_1}^{N_2})$} \\
    \hline
    \textsc{Smpnn}~\cite{ribalearning} & $\checkmark$ & $\checkmark$ & $\times$ & $\checkmark$ & {$O(\mathrm{max}(E_1, E_2, N_1 N_2))$} \\
    \textsc{EmbAvg} & $\checkmark$ & $\checkmark$ & $\times$ & $\checkmark$ & {$O(\mathrm{max}(E_1, E_2))$} \\
    \textsc{GCNMean}~\cite{kipf2016semi} & $\checkmark$ & $\checkmark$ & $\times$ & $\checkmark$ & {$O(\mathrm{max}(E_1, E_2))$} \\
    \textsc{GCNMax}~\cite{kipf2016semi} & $\checkmark$ & $\checkmark$ & $\times$ & $\checkmark$ & {$O(\mathrm{max}(E_1, E_2))$} \\
    \textsc{SimGNN}~\cite{bai2018graph} & $\checkmark$ & $\checkmark$ & $\times$ & $\checkmark$ & {$O(\mathrm{max}({N_1},{N_2})^2)$} \\
    \textsc{GMN}~\cite{li2019graph} & $\checkmark$ & $\checkmark$ & $\times$ & $\checkmark$ & {$O(\mathrm{max}({N_1},{N_2})^2)$} \\ \hline
    \textsc{\model} & $\checkmark$ & $\checkmark$ & $\times$ & $\checkmark$ & {$O(\mathrm{max}({N_1},{N_2})^2)$} \\
    \hline
\end{tabular}
\centering
\label{table:baseline_summary}
\end{table*}

In this section, we make the effort to address the following question: Why does \model perform consistently the best under most settings?

The result tables on four datasets under two metrics can be found in the main body of the paper.

\subsection{Approximate GED algorithms} As pointed out in Section~\ref{subsec-bgm}, \model learns from the ground-truth instead of directly computing the graph similarity. \model even outperforms some of the solvers that generate the ground-truth, e.g. \textsc{Hungarian} and \textsc{VJ} on \textsc{Ptc}, suggesting that \textsc{Hungarian} and \textsc{VJ} perform poor and \model instead learns mostly from \textsc{Beam} on this dataset.  

\subsection{\textsc{Smpnn}} It computes all pairwise node embedding similarity scores just like \model. However, it simply sums up the largest similarity score in each row and column of the similarity matrix. 
In contrast, \model is equipped with carefully designed CNN kernels, resulting in more trainable components, and uses multiple similarity ``images'' corresponding to comparing two graphs at different scales, feeding richer information to the CNN kernels.
The poor performance of \textsc{Smpnn} shows that care must be taken when utilizing node-node similarity scores in order to achieve good computation accuracy. 

\subsection{\textsc{EmbAvg}, \textsc{GCNMean} and \textsc{GCNMax}} They calculate the inner product of two graph-level embeddings as the predicted similarity score. \textsc{EmbAvg} performs unweighted average of node embeddings, while \textsc{GCNMean} and \textsc{GCNMax} adopt graph coarsening~\cite{dhillon2007weighted,defferrard2016convolutional} with mean and max pooling respectively, to obtain the graph-level embeddings. 
From both tables, their relatively poor performance could be attributed to the fact that the graph-level embeddings may be too coarse to capture fine-grained node-level interaction information. In other words, simply taking the inner products of graph-level embeddings may cause insufficient graph interaction, which is pivotal for detailed graph comparison. Interaction at finer granularity seems especially important since the definitions of the GED and MCS metrics involve node and edge matching. 

\subsection{\textsc{SimGNN}} This recent work compares two graphs via a two-level strategy: At the graph level, it uses a Neural Tensor Network~\cite{socher2013reasoning} module to interact two graph-level embeddings obtained by a node attention mechanism; at the node level, it uses the histogram features extracted from the node-node similarity scores.
However, the adoption of histogram function has the disadvantage of being not differentiable, resulting in sub-optimal usage of node-node similarity scores. What is worse, the histogram features completely ignore the node position information. In contrast, our model \model uses a node ordering technique to keep the position information of all graph nodes.

\subsection{\textsc{GMN}} 

To show that our proposed method \model is flexible enough to use any node embedding techniques, we conduct the following experiments: We replace the three GCN layers used in \model by three GMN's node embedding layers, where each layer performs a cross-graph attention mechanism to capture the structural difference for the two graphs at different scales. The key difference between this new \model (denoted as \model-w/\textsc{GMN}) and \textsc{GMN} is that, \model-w/\textsc{GMN} transforms the node embeddings generated by \textsc{GMN} to similarity scores, while \textsc{GMN} performs graph-level embeddings using the node embeddings directly.

As shown in Table~\ref{table:aids_with_gmn}, although \textsc{GMN} outperforms \model on the MCS metric on \textsc{Aids}, our model combined with the node embeddings by \textsc{GMN} performs better than either model alone, suggesting that the proposed technique indeed brings performance gain. 
\section{Theoretical Time Complexity}



The time complexity of generating multi-scale node embeddings for a graph is $O(|E|)$~\cite{kipf2016semi}, where $|E|$ is its number of edges. The similarity matrix generation has time complexity $O(\mathrm{max}({N_1},{N_2}))^2)$, including the BFS ordering and matrix multiplication. Note that we can take advantage of GPU acceleration for the dense matrix multiplication. The overall time complexity of \model for inference/testing is therefore $O(\mathrm{max}({N_1},{N_2}))^2)$. It is worth noting that the GED and MCS problems are NP-hard, and the proposed method has lower complexity than many existing approximate algorithms. Specifically, \model, \textsc{SimGNN}, and \textsc{GMN} have the same time complexity.




\section{Scalability Study}

\begin{table*}
\scriptsize
\centering
\vspace{-0.05in}
\caption{Results on \textsc{Reddit12k} under both the GED and MCS metrics. \textsc{A*} fails to compute most GEDs within a 5-minute limit (thus denoted as $-$). Instead, the minimum GED returned by \textsc{Beam}, \textsc{Hungarian}, and \textsc{VJ} for each pair is used as the ground-truth GED (labeled with superscript $*$). \textsc{Mcsplit} provides ground-truth MCS results, labeled with superscript $*$. The mse is in $10^{-3}$. The time is wall time in msec.}
  \begin{tabular}{lllllllll}
    \toprule
    \multirow{3}{*}{\textbf{Method}} &
      \multicolumn{4}{c}{\textbf{\textsc{Reddit12k} (GED)}} &
      \multicolumn{4}{c}{\textbf{\textsc{Reddit12k} (MCS)}} \\
      & \textbf{mse} & \textbf{$\rho$} & \textbf{p@10} & \textbf{time} & \textbf{mse} & \textbf{$\rho$} & \textbf{p@10} & \textbf{time}\\
      \midrule
     
    \textsc{A*} & $-$ & $-$ & $-$ & $-$ & $-$ & $-$ & $-$ & $-$ \\
    \textsc{Beam} & $15.153^{*}$ & $0.942^{*}$ & $0.983^{*}$ & $158.852^{*}$ & $-$ & $-$ & $-$ & $-$ \\
    \textsc{Hungarian} & $114.361^{*}$ & $0.743^{*}$ & $0.208^{*}$ & $31.068^{*}$ & $-$ & $-$ & $-$ & $-$ \\
    \textsc{VJ} & $116.496^{*}$ & $0.741^{*}$ & $0.258^{*}$ & $37.007^{*}$ & $-$ & $-$ & $-$ & $-$ \\
    \textsc{HED} & $510.073$ & $0.750$ & $0.392$ & $94.494$ & $-$ & $-$ & $-$ & $-$ \\
    \textsc{Mcsplit} & $-$ & $-$ & $-$ & $-$ & $0.000^{*}$ & $1.000^{*}$ & $1.000^{*}$ & $8882.773^{*}$ \\ \hline
    \textsc{Smpnn} & $35.921$ & $0.002$ & $0.001$ & $14.609$ & $41.580$ & $0.010$ & $0.005$ & $14.523$ \\
    \textsc{EmbAvg} & $17.167$ & $0.271$ & $0.017$ & $\textbf{7.154}$ & $10.657$ & $0.471$ & $0.142$ & $\textbf{7.933}$ \\
    \textsc{GCNMean} & $19.842$ & $0.084$ & $0.002$ & $8.009$ & $13.999$ & $0.231$ & $0.002$ & $8.194$ \\
    \textsc{GCNMax} & $19.070$ & $0.003$ & $0.002$ & $8.591$ & $13.865$ & $0.188$ & $0.075$ & $8.001$ \\
    \textsc{SimGNN} & $3.822$ & $0.763$ & $0.300$ & $16.720$ & $6.533$ & $0.737$ & $0.317$ & $15.431$ \\
    \textsc{GMN} & $14.702$ & $0.458$ & $0.108$ & $17.192$ & $7.053$ & $0.747$ & $0.308$ & $17.993$ \\
    \textsc{\model} & $3.511$ & $0.794$ & $0.333$ & $26.490$ & $6.164$ & $0.793$ & $0.350$ & $27.301$ \\
    \textsc{\model-w/\textsc{GMN}} & $\textbf{3.479}$ & $\textbf{0.842}$ & $\textbf{0.397}$ & $38.651$ & $\textbf{6.157}$ & $\textbf{0.850}$ & $\textbf{0.483}$ & $38.213$ \\ \hline
  \end{tabular}
\centering
\label{table:reddit_with_gmn}
\end{table*}

To study the scalability of \model and compare with baselines, we run all the methods on \textsc{Reddit12k}~\cite{yanardag2015deep}, which contains 11929 graphs each corresponding to an online discussion thread where nodes (unlabeled) represent users, and an edge represents the fact that one of the two users responded to the comment of the other user. This is so far the largest real-world graph dataset in terms of the number of graphs (11929), average graph size (386.2), and the maximum graph size (3760) widely used by research works on graph classification~\cite{ying2018hierarchical,verma2018graph}.

As shown in Table~\ref{table:reddit_with_gmn}, \textsc{\model-w/\textsc{GMN}} achieves the best performance in terms of effectiveness. Although slower than simpler neural network methods like \textsc{EmbAvg}, the proposed method is still faster than most approximate GED and MCS computation algorithms (i.e. \textsc{Beam}, \textsc{Hungarian}, \textsc{VJ}, \textsc{HED}, and \textsc{Mcsplit}), and better/more effective than the simpler methods. It is noteworthy that \textsc{GCNMean} and \textsc{GCNMax} perform worse than \textsc{EmbAvg}, a phenomenon not observed for the other four datasets shown in the main text. This is likely due to the fact that on average the graphs in \textsc{Reddit12k} are larger than the other four datasets, and using the coarsening algorithm with GCN as proposed in \cite{defferrard2016convolutional} would not be very effective on the larger graphs. It would be interesting in future to investigate more powerful techniques to extract hierarchical features for large graphs such as \textsc{DiffPool}~\cite{ying2018hierarchical} and adopt in graph similarity tasks.
\section{Parameter Sensitivity}

\begin{figure}[h]
    \centering
    {{\includegraphics[width=0.22\textwidth]{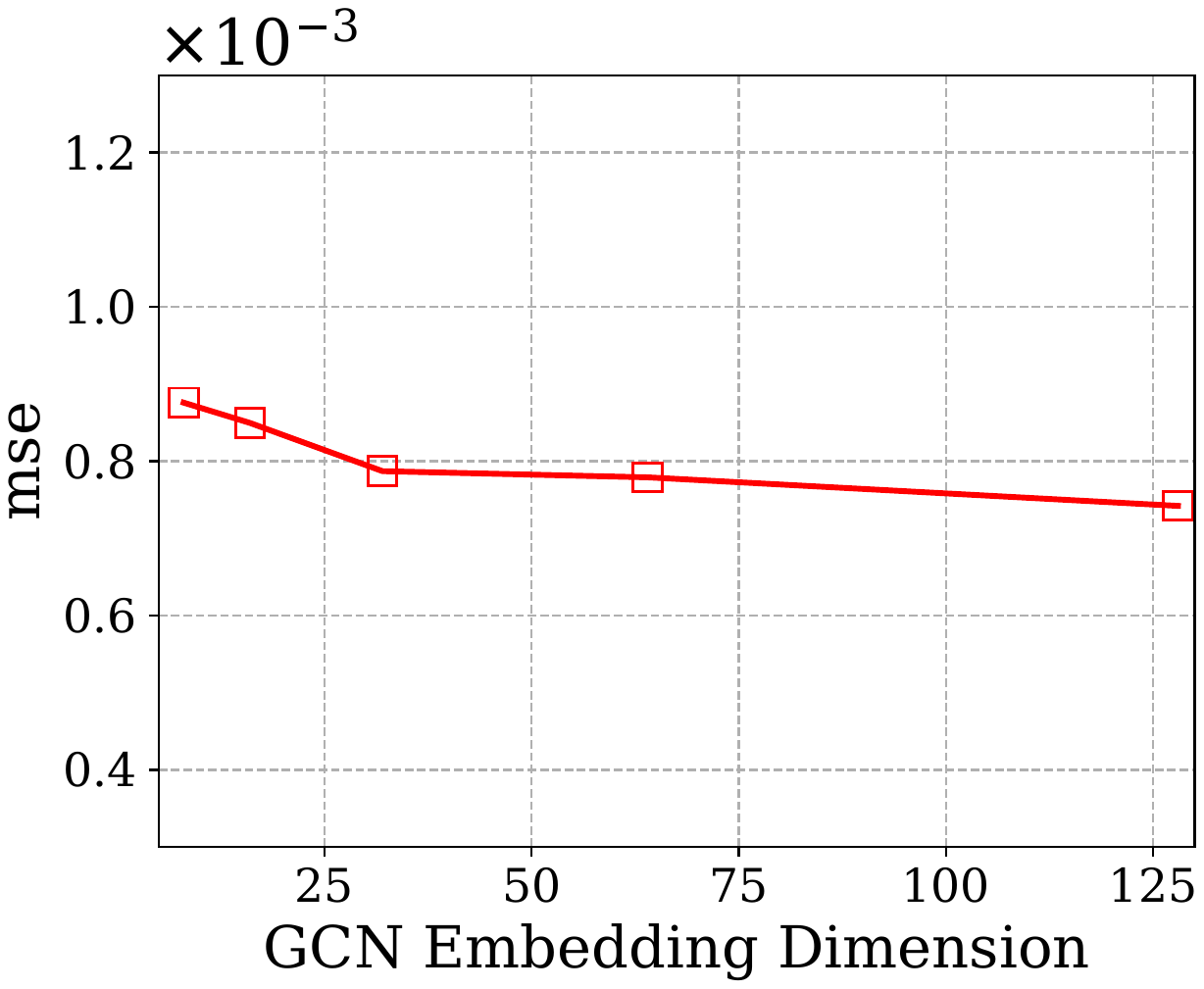}}}
    {{\includegraphics[width=0.22\textwidth]{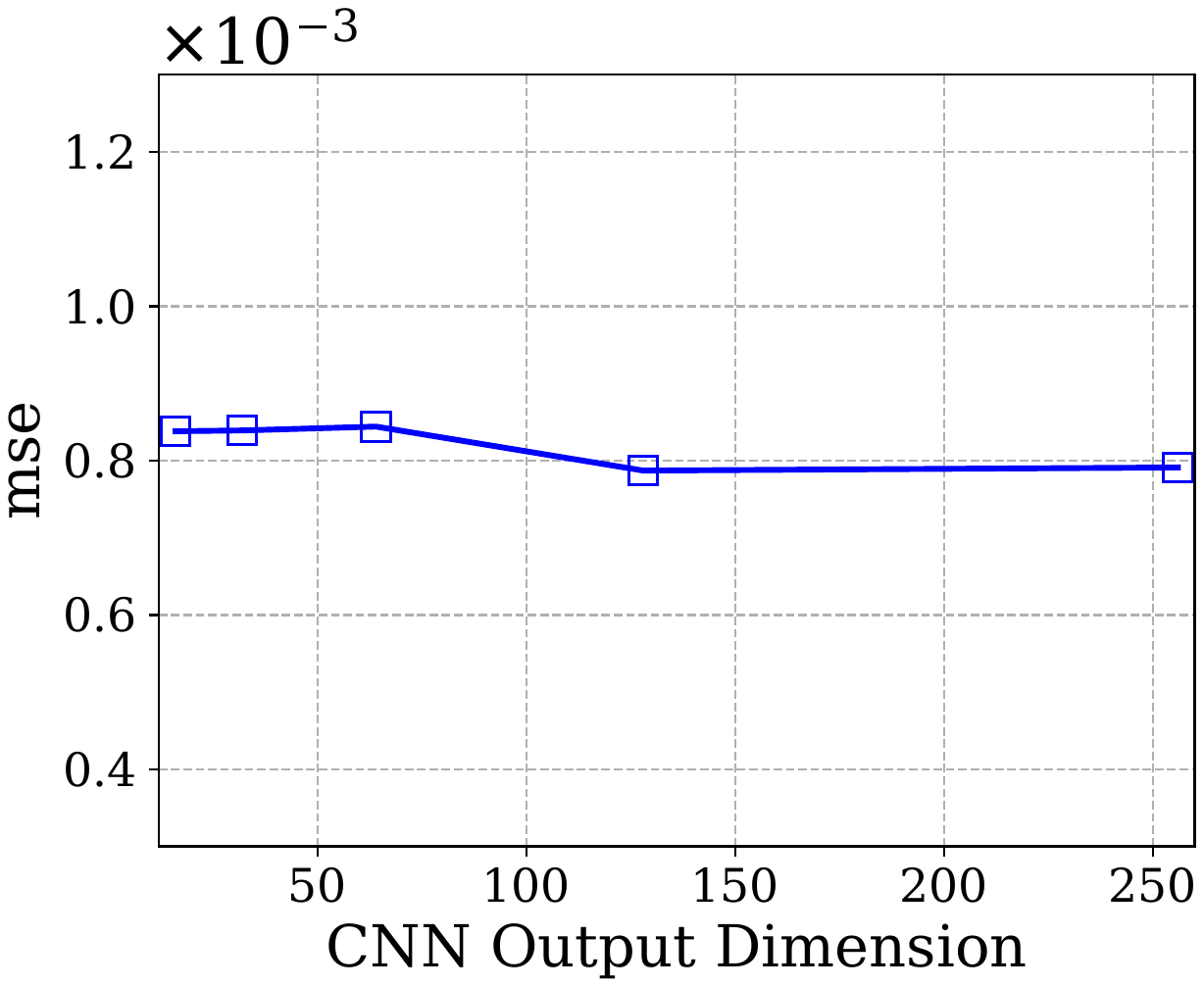}}}
    {{\includegraphics[width=0.22\textwidth]{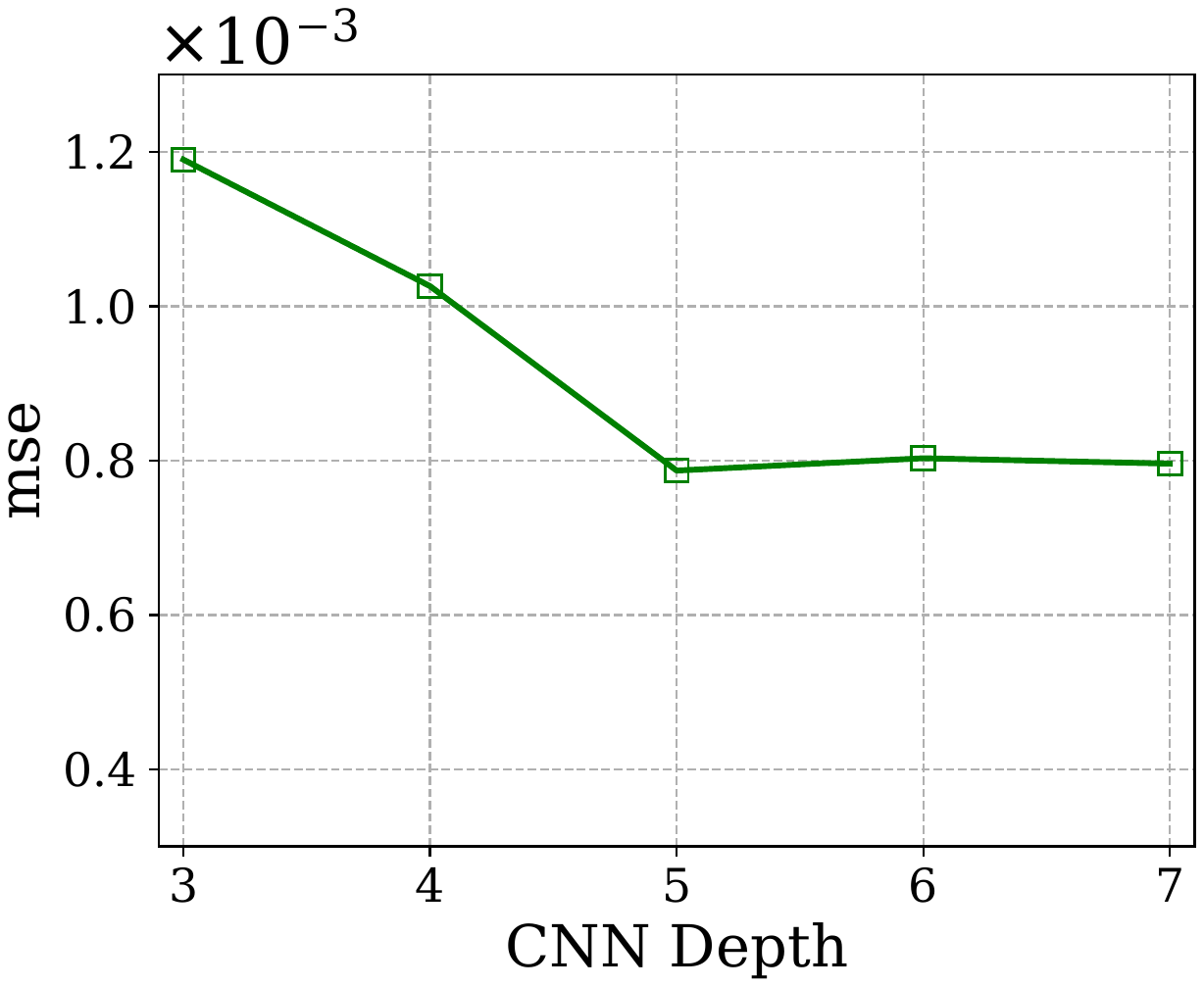}}}
    {{\includegraphics[width=0.22\textwidth]{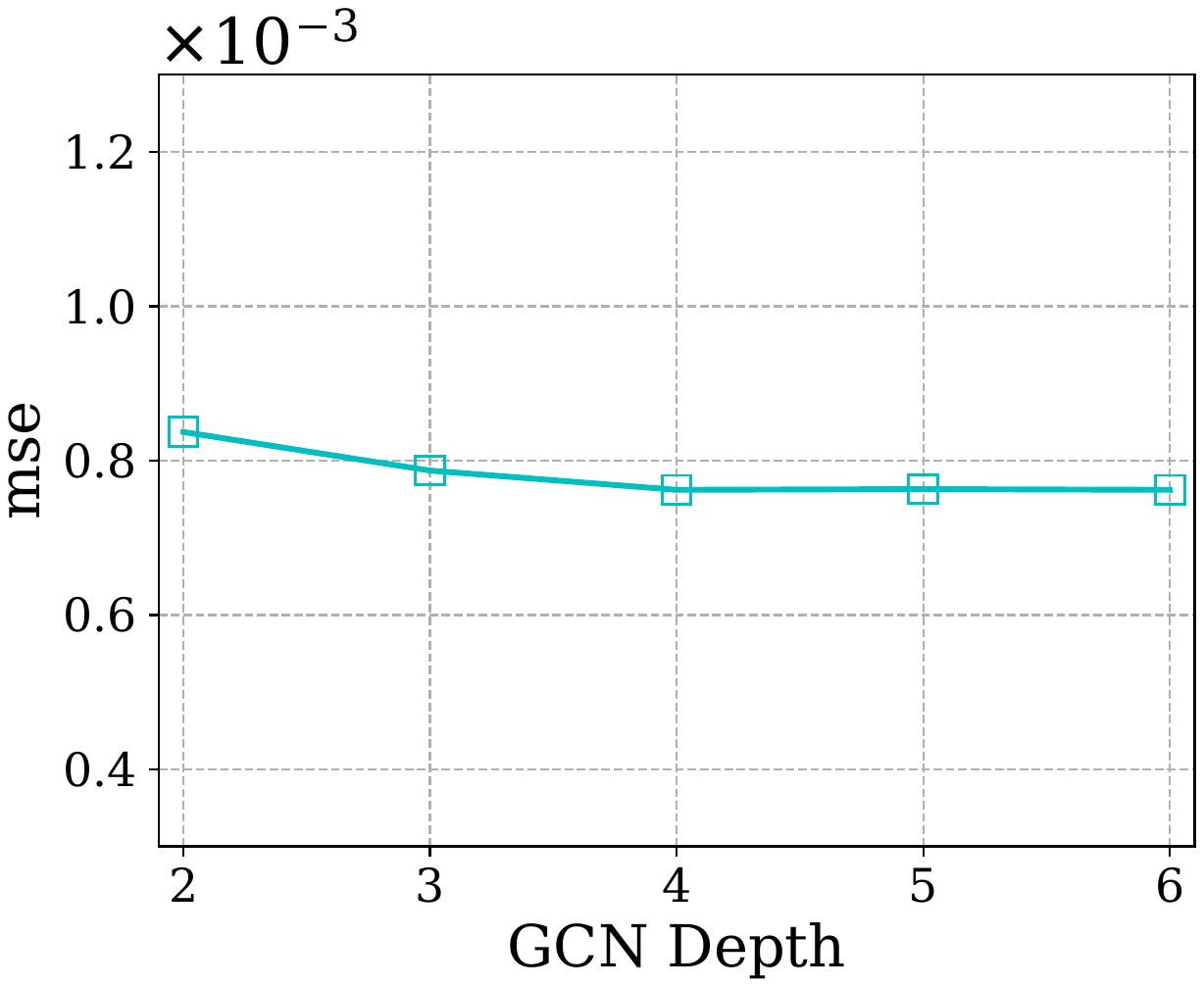}}}
    \caption{Mean squared error on the \textsc{Aids} dataset w.r.t. the dimensions of GCN output, the dimensions of CNN output, the number of CNN and GCN layers.}
    \label{fig:param}
\end{figure}

We evaluate how the dimensions of the GCN node embedding, the dimensions of the CNN output, the number of CNN layers in each channel, and the number of GCN layers can affect the results. We report the mean squared error on \textsc{Aids}. 

As can be seen in Fig.~\ref{fig:param}, the performance becomes marginally better if larger dimensions are used. Fig.~\ref{fig:param} shows that the more the CNN layers, the better the performance. However, after five layers, the performance does not change much. Overall, the performance is more sensitive to CNN depth than to the dimensions of GCN embedding, CNN output, and the number of GCNs.

Note that the performance of \model does not change much if additional GCN layers are used. It makes sense since more GCN layers would cause information at larger scales to be captured by the CNN kernels, and after certain number of scales, the model should be able to ``see'' enough structural and attribute difference between two graphs.

\section{Related Tasks}

\subsection{Related Task 1: Graph Database Search}

In a recent work~\cite{liang2017similarity} on graph database search 
the authors propose a filtering-verification approach, aiming to reduce the total amount of exact GED computations for a query to a tractable degree, through a series of database indexing techniques and pruning strategies. \cite{liang2017similarity} also uses the \textsc{Aids} dataset (albeit the full version with more graphs), along with one additional real graph dataset (database) with 32.6 nodes on average.

Although their goals and approach are different from ours, both works involve the computation of GED, which is NP-hard in nature, to be computed exactly. Therefore, the exact GED computation is intractable on  large graphs.
In this work, we choose an alternative strategy to deal with relatively large graphs to obtain ground-truth to train and evaluate our model, which will be described in Section~\ref{subsec-data-preproc}.

In the future, it is promising to explore possibilities of combining the proposed framework with techniques proposed by the graph database community to tackle the database search problem. 

\subsection{Related Task 2: Graph Alignment}

Graph alignment and graph similarity computation are both under the general umbrella of graph matching. However, the former is more about node-level alignments/mapping, while the latter is more about graph-level similarity computation. 

To better understand the connection and distinction, we take a recent work on graph alignment~\cite{heimann2018regal} as an example. \cite{heimann2018regal} describes graph alignment as ``finding corresponding nodes in different networks''. Its ground-truth node correspondence is generated by permuting the node ids of a graph $\mathcal{G}$ to obtain an isomorphic $\mathcal{G}'$ followed by randomly removing edges of $\mathcal{G}'$ without disconnecting any nodes. 
In contrast, \model learns from any general graph similarity metric, e.g. GED and MCS, which can take any type of graph transforming operations, e.g. node deletion. \model by design is more flexible and is able to learn from any graph similarity metric as long as the true similarity scores under the desired metric are provided to train the model. 

In future, it would be interesting to see how \model can be extended to tackle the task of graph alignment, requiring the output of finer comparison at the node level. 


\end{document}